%% file: acl_latex.tex
\documentclass[11pt]{article}

\usepackage[final]{acl}

\usepackage{times}
\usepackage{latexsym}
\usepackage{enumitem}

\usepackage[T1]{fontenc}

\usepackage[utf8]{inputenc}
\usepackage{xspace}

\usepackage{microtype}
\usepackage{multicol}
\usepackage{amssymb}

\usepackage{inconsolata}
\usepackage[most]{tcolorbox}
\newtcblisting{jsonbox}{
  breakable,
  listing only,
  colback=gray!12,
  colframe=black,
  boxrule=0.8pt,
  arc=2pt,
  left=8pt,
  right=8pt,
  top=8pt,
  bottom=8pt,
  listing options={
    basicstyle=\ttfamily\small\linespread{1.1}\selectfont,
    columns=fullflexible,
    keepspaces=true,
    breaklines=true,
    breakindent=0pt, 
    showstringspaces=false,
    aboveskip=0pt,
    belowskip=0pt
  }
}
\usepackage{dsfont}

\usepackage{graphicx}
\usepackage{amsmath}
\usepackage{subcaption}
\usepackage{booktabs}
\usepackage[table]{xcolor}
\usepackage{pifont}
\usepackage{tabularx}
\usepackage{longtable}

\definecolor{DeepGreen}{RGB}{0,160,80}

\definecolor{mygreen}{HTML}{B9FFBF}
\definecolor{mylightgreen}{HTML}{D9F0D9}
\definecolor{purple1}{RGB}{126, 107, 196}
\definecolor{purple2}{RGB}{199, 158, 207}
\definecolor{purple3}{RGB}{214, 200, 255}
\definecolor{purple4}{RGB}{254, 240, 255}
\title{POINTS-Seeker: An Open Recipe for Multimodal Search Agents with Visual Memory Management}

\author{
  \hspace{-0.25cm}\textbf{Yikun Liu$^{1,2\star}$, Yuan Liu$^{3\star}$, Haicheng Wang$^{3}$, Zhongyin Zhao$^{3}$, Le Tian$^{3}$,} \\
  \hspace{-0.25cm}\textbf{Xiao Zhou$^{3}$, Jiangchao Yao$^{2}$, Yanfeng Wang$^{1}$, Weidi Xie$^{1}$} \\
  \hspace{-0.25cm}$^{1}$School of Artificial Intelligence, Shanghai Jiao Tong University \\
  \hspace{-0.25cm}$^{2}$CMIC, Shanghai Jiao Tong University \quad
  $^{3}$WeChat AI, Tencent \\
  \hspace{-0.25cm}\small $^\star$These authors contributed equally to this work.
}

\begin{document}
\maketitle

\input{sections/0_abstract}
\input{sections/1_intro}

\input{sections/2_related-work}
\input{sections/3_method}
\input{sections/4_experiments}

\input{sections/5_conclusion}
\input{sections/limitation}


\bibliography{custom}


\input{sections/X_supp}

\end{document}

%% file: sections/0_abstract.tex
\begin{abstract}

Large Multimodal Models~(LMMs) excel at visual perception but struggle with real-time, knowledge-intensive queries due to their reliance on static parametric knowledge. While multimodal search agents offer a promising solution, developing them from vanilla LMMs presents two major challenges: the lack of open recipes to cultivate search agency from scratch, and the severe context explosion (attention dilution) that occurs during long-horizon multi-turn investigations. In this paper, we address both challenges by developing a \textbf{native} multimodal search agent. First, we introduce an open training recipe featuring \textit{Agentic Seeding}, a formative training stage that bootstraps tool-use and planning abilities directly from a non-agentic foundation. Second, to overcome the context bottleneck, we propose \textbf{V-Fold}, an adaptive memory management mechanism. V-Fold retains recent interactions as high-fidelity text while ``folding'' stale historical context into the visual space via rendering, exploiting the model's cross-modal alignment to preserve raw evidence without text token redundancy. Combining these innovations, we present \textbf{POINTS-Seeker-8B}, which achieves state-of-the-art performance among models of comparable scale across six multimodal search benchmarks.


\end{abstract}


%% file: sections/1_intro.tex
\section{Introduction}
\label{sec:intro}

Large Multimodal Models~(LMMs)~\cite{bai2025qwen3, liu2025points, liu2024points1, guo2025seed1, team2025kimi} have demonstrated impressive proficiency in static perception and reasoning. However, their reliance on static parametric knowledge makes them prone to hallucinations when faced with real-time, knowledge-intensive queries. To address this limitation, recent work has shifted toward agentic search, enabling LMMs to autonomously gather and synthesize web evidence for complex visual question answering~(VQA) tasks~\cite{liu2025visual, yao2026mmdeepresearch, wu2025mmsearch, narayan2025deepmmsearch}.

Despite these advancements, developing a robust multimodal search agent remains challenging on two fronts. \textbf{First}, prevailing search agents~\cite{wu2025mmsearch, huang2026vision, narayan2025deepmmsearch, zhang2025skywork} are typically built upon foundations that have already undergone extensive proprietary agentic mid-training. This paradigm formulates multimodal search as a downstream specialization, leaving a critical gap in understanding how to systematically cultivate search agency from a vanilla, non-agentic LMM. \textbf{Second}, as agents acquire the ability to conduct multi-turn investigations, they inevitably suffer from context explosion. The accumulation of lengthy web snippets inflates the interaction history, leading to severe ``attention dilution'' (akin to the ``Lost-in-the-Middle'' effect~\cite{liu2024lost}). Consequently, the model often fails to locate ground-truth evidence even when it is present in its context~(Fig.~\ref{fig:history_hit}), rendering complex reasoning prone to failure.

To address these coupled challenges, we introduce an open training recipe and a novel visual memory management mechanism to build a \textit{native} multimodal search agent from scratch. To tackle the first challenge, we design a staged training curriculum powered by an automated data curation pipeline. A key innovation in this recipe is \textbf{Agentic Seeding}---a formative stage immediately following pre-training. By exposing the model to dense, long-horizon reasoning and ReAct-style~\cite{yao2022react} trajectories, Agentic Seeding bootstraps the model's intrinsic abilities in tool use, planning, and evidence seeking. Subsequent Supervised Fine-Tuning~(SFT) and Reinforcement Learning~(RL) stages further refine these behaviors, allowing us to study the complete evolution of a vanilla LMM into a capable search agent.

\begin{figure*}[t]
  \centering
  \includegraphics[width=\textwidth]{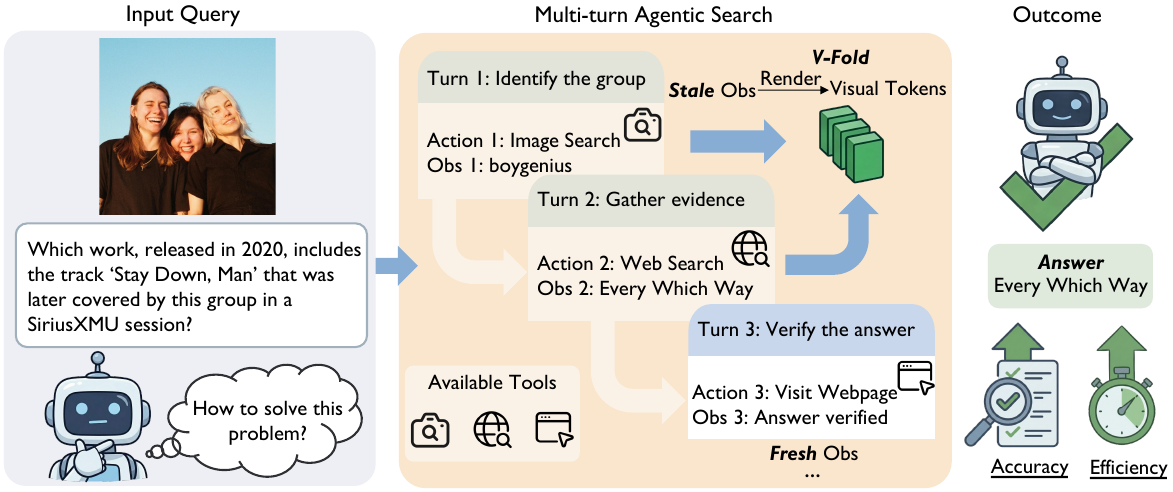} \\
  \vspace{-4pt}
  \caption{\textbf{Overview of POINTS-Seeker.} Our model adaptively interacts with external tools to solve complex, multi-hop VQA tasks. By leveraging V-Fold compression, stale history is rendered into compact visual tokens, effectively bypassing long-context performance degradation while achieving efficiency and reasoning fidelity.}
  \vspace{-1.5em}
 \label{fig:teaser}
\end{figure*}

To tackle the second challenge of context explosion, we propose \textbf{V-Fold}, an adaptive, history-aware folding mechanism that uniquely exploits the multimodal nature of the agent. Unlike conventional text-only strategies such as naive truncation or summarization~\cite{yu2025memagent, liu2025deepseek} which risk discarding critical cues, V-Fold shifts the information burden from textual processing to visual perception. Once the interaction history reaches a complexity threshold, V-Fold adopts a sliding-window strategy: recent interactions are retained as high-fidelity text for immediate reasoning, while stale historical observations are ``folded'' into the visual space by rendering them into images. This dual-representation ensures the model can still ``see'' the raw historical evidence via its native vision-language alignment without being overwhelmed by textual token redundancy. Notably, V-Fold is highly data-efficient; by injecting merely 5k rendered trajectories during SFT, the model internalizes this cross-modal folding capability, supporting robust out-of-the-box deployment.

By integrating this training recipe with V-Fold, we present \textbf{POINTS-Seeker-8B}. Extensive experiments across six benchmarks demonstrate that our model substantially outperforms existing agents based on post-hoc fine-tuning, effectively mitigating the bottleneck in long-horizon interactions.

In summary, our main contributions are as follows: (i) we introduce an open, systematic training recipe to cultivate multimodal search agency from a vanilla LMM, highlighting \textit{Agentic Seeding} as a crucial formative stage for bootstrapping tool-use and planning abilities; (ii) we propose \textit{V-Fold}, a novel visual memory management mechanism that adaptively renders stale textual history into compact visual tokens, effectively alleviating attention dilution in long-horizon search without losing raw evidence; (iii) we develop POINTS-Seeker-8B, which achieves state-of-the-art performance among models of comparable scale across six challenging multimodal search benchmarks.

%% file: sections/2_related-work.tex
\section{Related Work}
\label{sec:related_work}

\noindent \textbf{Retrieval-augmented generation.}
Retrieval-augmented generation (RAG) grounds LLM responses in retrieved information, mitigating the static nature of model parameters~\cite{karpukhin2020dense, lewis2020retrieval}. This paradigm has been widely extended to multimodal settings, especially knowledge-intensive VQA, where external text, images, and structured multimodal data provide evidence for complex reasoning~\cite{yu2024visrag, liu2025lamra, chen2024mllm, chen2022murag, hu2023reveal, yan2024echosight, mensink2023encyclopedic, shi2026weaver}. However, conventional RAG methods typically rely on static knowledge bases and decouple retrieval from generation, limiting end-to-end optimization. Recent work thus shifts toward autonomous web-search agents that interact with real-time web environments~\cite{chen2026opensearch, zheng2025deepresearcher, li2026hypereyesdualgrainedefficiencyawarereinforcement, redsearcher2026}. In multimodal domain, methods such as MMSearch-R1, DeepMMSearch-R1, and related systems~\cite{wu2025mmsearch, narayan2025deepmmsearch, wang2025vrag, du2026longhorizonagenticmultimodalsearch, huang2026onpolicydataevolution} further improve search agents via end-to-end RL. In contrast, this paper studies how to systematically cultivate search agency from a vanilla, non-agentic multimodal model through an open recipe and visual memory management.

\smallskip \noindent \textbf{Agent memory.} Continuous interactions often cause context explosion, motivating various memory management strategies~\cite{chhikara2025mem0, wang2025mem, yan2025memory}. For example, MemAgent~\cite{yu2025memagent} proposes maintaining a fixed-length memory that is proactively updated by the model. AgentFold~\cite{ye2025agentfold} introduces a folding mechanism to intelligently fold segments of context during task execution. Recently, cross-modal compression~\cite{shi2026memocr, feng2026agentocr} has emerged as a potent alternative. DeepSeek-OCR~\cite{wei2025deepseek} proposes that images can serve as an effective medium for compression. Similarly, Glyph~\cite{cheng2025glyph} advances a related idea by transforming the long-text modeling problem into a multiple image–text modeling paradigm. Furthermore, AgentOCR~\cite{feng2026agentocr} proposes a self-compression mechanism, in which the model determines the compression ratio. However, memory management in multimodal search remains under-explored. In this paper, we study how history-aware memory folding can preserve critical evidence while mitigating context overload in long-horizon multimodal search.

%% file: sections/3_method.tex
\section{Method}
\label{sec:method}

This section starts by formulating the problem in Sec.~\ref{subsec:problem}, followed by the description of our open training recipe and data curation pipeline in Sec.~\ref{subsec:recipe}. Finally, Sec.~\ref{subsec:compression} elaborates on V-Fold, our adaptive visual compression mechanism.

\subsection{Problem Formulation}
\label{subsec:problem}
We consider the task of tackling complex visual question answering in a real-world web environment. Given a query~($q$), which typically consists of a textual question and an image, the policy model~($\mathcal{M}_{\theta}$) determines the action at each interaction turn. Specifically, it may either invoke an external tool to acquire additional information or generate the final answer directly. The interaction process terminates once the final answer is produced. We define the core components as follows:

\begin{itemize}[leftmargin=*, topsep=0.1ex, itemsep=-0.1ex]
    \item \textbf{Action~($a_t$)}. The atomic operation executed at step~$t$. An action may either invoke an external tool~(as listed in Sec.~\ref {sec:exp_setup}) or generate the answer.
    
    \item \textbf{Observation~($o_t$)}. The result returned by the tool after executing $a_t$. Observation may include webpage URLs, titles, and snippets, among other information. Please refer to Sec.~\ref{sec:exp_setup} for details.
    

    \item \textbf{History~($\mathcal{H}_t$)}. The sequence of preceding events up to step~$t$:
    $\mathcal{H}_t=(q,a_1,o_1,\ldots,a_{t-1},o_{t-1})$. The policy model~($\mathcal{M}_{\theta}$) parameterizes an action distribution~($\pi_{\theta}$), from which the next action is sampled:
    $a_t \sim \pi_{\theta}(\cdot \mid \mathcal{H}_t)$.

    
    \item \textbf{Answer~($y$)}. The final response is generated upon collecting sufficient evidence or reaching the maximum interaction limit.
\end{itemize}

\subsection{Training Pipeline}
\label{subsec:recipe}



\begin{figure*}[t]
  \centering
  \includegraphics[width=\textwidth]{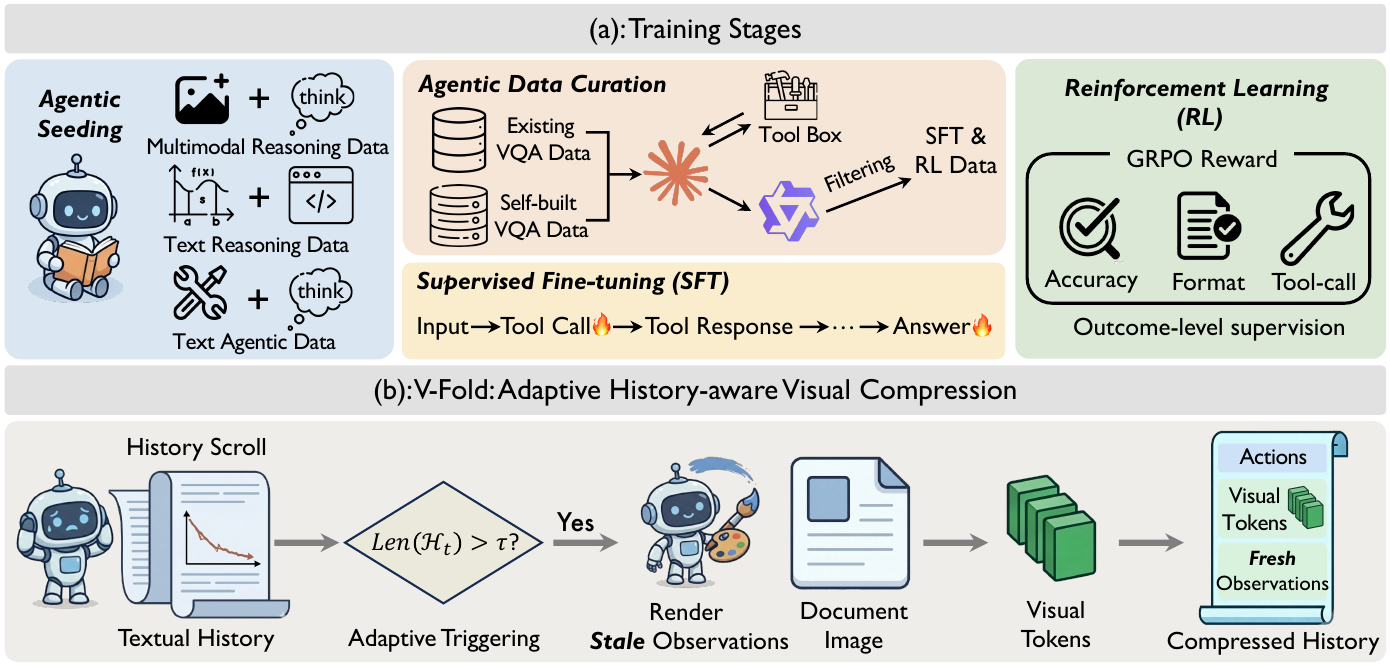} \\
  \vspace{-6pt}
  \caption{\textbf{The framework of our agentic search system.} \textbf{(a)} Training stages, including Agentic Seeding, SFT, and RL. \textbf{(b)} V-Fold, which adaptively compresses stale history into visual tokens for efficient long-context reasoning.}
  \vspace{-16pt}
 \label{fig:arch}
\end{figure*}

Starting from a vision-language model with standard visual-language pre-training (see supplementary material for details), we develop agentic search capabilities through a three-stage training pipeline, as illustrated in Fig.~\ref{fig:arch}. First, \textit{Agentic Seeding} injects general reasoning, planning, and tool-use 
priors into the model through mid-training. We then curate task-specific agentic search data for the subsequent stages. Second, SFT aligns the model with accurate tool invocation and response formatting using curated trajectories. Finally, RL further optimizes the model to improve correctness, reduce invalid tool calls, and encourage exploration.

\subsubsection{Stage I: Agentic Seeding}
\label{subsec:agentic_seeding}


Agentic Seeding is a mid-training stage designed to cultivate general agentic capabilities before task-specific search training. Rather than directly adapting the model to a particular tool environment, this stage strengthens its reasoning, planning, and preliminary tool-use priors. We train the model with the standard next-token prediction objective on a mixture of the following data:

\begin{itemize}[leftmargin=*, topsep=0.1ex, itemsep=-0.1ex]
    \item \textbf{Multi-modal reasoning data}: High-quality VQA data featuring dense reasoning traces to align visual perception with logical reasoning.
    \item \textbf{Text reasoning data}: Diverse logic-intensive datasets, primarily centering on mathematical and coding corpora, that instill system-2 thinking. This curriculum provides the cognitive foundation for the model's internal world model, facilitating complex planning across modalities~\cite{tu2024mlan, wei2025open}.
    \item \textbf{Text agentic data with ReAct style}: ReAct-style textual agent data that foster fundamental tool-use logic and planning strategies. By exposing the model to these interaction patterns early on, we facilitate the seamless transfer of agentic priors from the textual to the multimodal domain~\cite{tu2024mlan, wei2025open}.
\end{itemize}

This stage yields a checkpoint with stronger reasoning and planning priors, which is then used as the initialization for the subsequent SFT stage.



\subsubsection{Agentic Search Data Curation}
\label{subsec:data}


 After Agentic Seeding, we construct task-specific agentic search data for the following SFT and RL stages. It produces two types of training resources: 
(i) high-quality tool-use trajectories for SFT; 
(ii) VQA instances with ground-truth answers for RL.


\vspace{2pt}
\noindent \textbf{VQA data construction.} We curate high-quality VQA data from search-oriented VQA datasets~({\em i.e.}, LiveVQA~\cite{fu2025seeking} and FVQA~\cite{wu2025mmsearch}), and by converting complex multi-hop QA datasets~\cite{xia2025open}. Our automated pipeline has three key steps: (i) \textbf{Entity extraction \& selection:} We use Claude to identify visually representable entities from questions. (ii) \textbf{Query de-biasing:} To ensure retrieval accuracy, we generate context-aware search queries ({\em e.g.}, appending ``New South Wales politician'' to ``Mike Baird'') to resolve potential ambiguities. (iii) \textbf{VQA reformulation:} The question is rewritten to implicitly reference the target entity as ``the person/object in the image.'' Finally, we pair the reformulated question with an image retrieved via a search engine to form a coherent, search-intensive VQA instance.

\smallskip \noindent \textbf{Trajectory generation and filtering.} Given the (\textit{image, question, answer}) triplets, we synthesize structured trajectories composed of sequential \texttt{<think>}, \texttt{<tool\_call>}, \texttt{<tool\_response>}, and \texttt{<answer>} segments. Specifically, Claude is employed to generate trajectories with tool-call annotations. Whenever a \texttt{<tool\_call>} is produced, the corresponding tool is executed, and its \texttt{<tool\_response>} is appended to the context for the next query, iteratively constructing the full trajectory. We then use \texttt{Qwen3-235B-A22B} as an LLM judge to compare the predicted answer against the ground truth, retaining only correct trajectories. This process yields 5,439 high-fidelity trajectories for SFT and 5,339 samples for RL training.

\begin{figure*}[t]
    \centering
    \begin{subfigure}[t]{0.55\textwidth}
        \centering
        \includegraphics[
            width=\linewidth,
            height=4cm,
            keepaspectratio
        ]{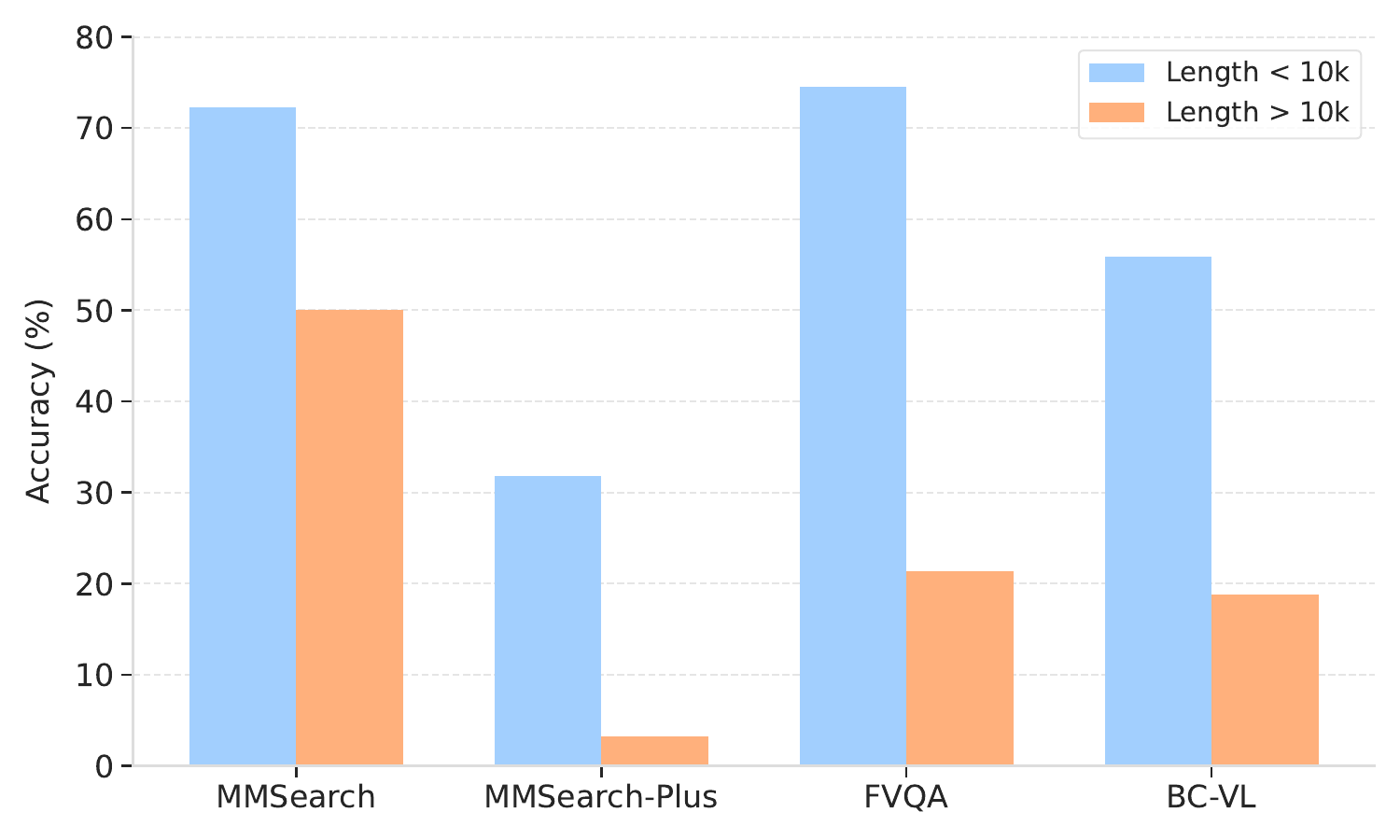}
        \caption{Comparison of model accuracy categorized by interaction length.}
        \label{fig:long_sequence}
    \end{subfigure}
    \hfill
    \begin{subfigure}[t]{0.43\textwidth}
        \centering
        \includegraphics[
            width=\linewidth,
            height=4cm,
            keepaspectratio
        ]{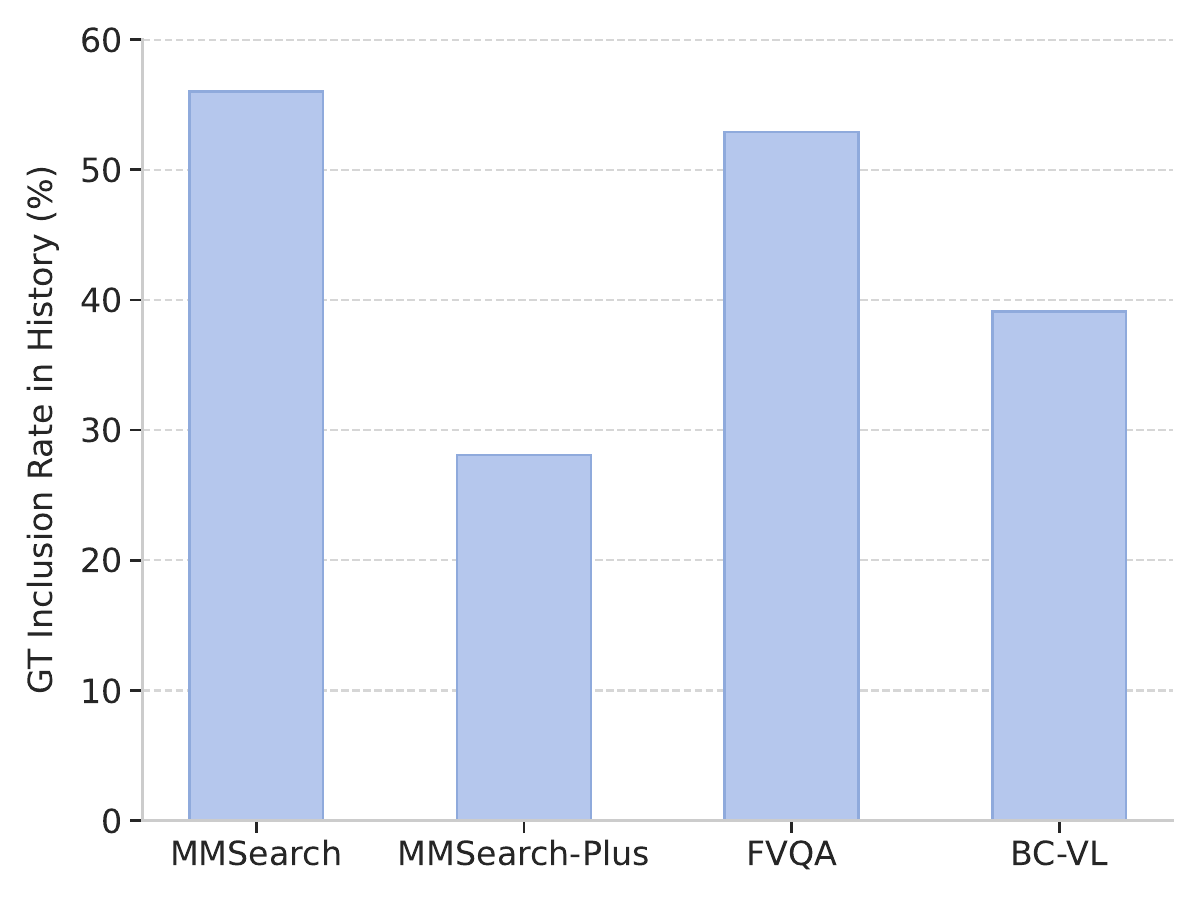}
        \caption{Ratio of history containing GT in failure cases.}
        \label{fig:history_hit}
    \end{subfigure}
    \caption{\textbf{Motivation for history-aware compression.} (a) Performance degradation in long-context scenarios. (b) The presence of GT in the failure case history indicates that compression must minimize historical information loss.}
    \label{fig:compression_motivation}
    \vspace{-1.5em}
\end{figure*}

\subsubsection{Stage II: Supervised Fine-Tuning}
\label{subsec:sft}

Starting from the Agentic-Seeding checkpoint, we perform supervised fine-tuning on the curated high-quality tool-use trajectories. 
This stage adapts the model's general reasoning and planning priors to the target agentic search environment. During the cold-start phase, we perform token-level supervision on the curated trajectories. For a training batch of $N$ samples, let $X^i = \{x_1^i, x_2^i, \dots, x_{L_i}^i\}$ denote the $i$-th training sample of length $L_i$, which comprises the query, reasoning thoughts, tool calls, tool responses, and the final answer. The training objective minimizes the negative log-likelihood:

\vspace{-2pt}
$$\mathcal{L}(\theta) = -\frac{1}{N} \sum_{i=1}^{N} \sum_{j=1}^{L_i} m_j^i \log P_\theta \left( x_j^i \mid x_{<j}^i \right)$$
where $m_j^i \in \{0, 1\}$ is a binary learning mask. We set $m_j^i=1$ for model-generated tokens, i.e., those in \texttt{<think>}, \texttt{<tool\_call>}, or \texttt{<answer>} blocks, and $m_j^i=0$ for the input query $q$ and tool-returned \texttt{<tool\_response>} tokens, ensuring the model learns only its own reasoning and actions.

\subsubsection{Stage III: Reinforcement Learning}
\label{subsec:rl}

After SFT, the model gains basic tool-use abilities but may still produce redundant tool calls or inconsistent formats. We thus adopt a tool-augmented variant of GRPO~\cite{guo2025deepseek} to avoid noisy trajectories and encourage exploration.

\vspace{2pt}
\noindent \textbf{Rollouts.} The policy model iteratively calls tools via pre-defined schemas, appending tool responses to the context. This process halts upon generating an answer or reaching a turn limit (e.g., 10). 

\vspace{2pt}
\noindent \textbf{Reward.} The RL training employs a multi-component reward function to optimize both reasoning quality and interaction integrity:
\begin{itemize}[leftmargin=*, topsep=0.1ex, itemsep=-0.1ex]
\item \textbf{Accuracy ($S_{\text{acc}}$)}: A correctness score $\{0, 1\}$ assigned by \texttt{Qwen3-235B-A22B} by comparing the model's output with the ground truth.
\item \textbf{Format ($S_{\text{format}}$)}: A binary reward ensuring reasoning and final answers are correctly enclosed within \texttt{<think>} and \texttt{<answer>} tags, respectively.
\item \textbf{Tool-call ($S_{\text{tool}}$)}: A binary score verifying if tool invocations adhere to the predefined schema and are properly wrapped in \texttt{<tool\_call>} tags.
\end{itemize}

The total reward is defined as \( R_{\text{total}} = \lambda_{\text{acc}}S_{\text{acc}} + \lambda_{\text{format}}S_{\text{format}} + \lambda_{\text{tool}}S_{\text{tool}} \), where $\lambda_{\text{acc}}$, $\lambda_{\text{format}}$ and $\lambda_{\text{tool}}$ denote the corresponding reward weights.

\subsection{V-Fold: Adaptive Visual Compression}
\label{subsec:compression}



V-Fold addresses long-horizon agentic search by adaptively compressing stale textual observations into compact visual representations. 
This design is motivated by two empirical findings: as shown in Fig.~\ref{fig:long_sequence}, model performance drops as interaction length increases; meanwhile, Fig.~\ref{fig:history_hit} shows that the ground-truth evidence is often already contained in the histories of failure cases. 
These observations suggest that failures are largely due to attention dilution over long contexts rather than missing evidence. 
Thus, instead of truncating or aggressively summarizing historical observations, V-Fold preserves raw evidence by rendering stale observations as images, while retaining the most recent interactions in text to maintain local reasoning continuity.

\input{tables/sota}
Specifically, the history at step $t$ is denoted as $\mathcal{H}_t = (q, a_1, o_1, \dots, a_{t-1}, o_{t-1})$. We define a triggering criterion $\mathds{1}_{\text{compress}}$ based on a predefined token threshold $\tau$:
\begin{equation*}
    \mathds{1}_{\text{compress}} = 
    \begin{cases} 
    1, & \text{if } \text{Len}(\mathcal{H}_t) > \tau \\ 
    0, & \text{otherwise}
    \end{cases}
\end{equation*}
When $\mathds{1}_{\text{compress}} = 1$, the compression mechanism is triggered. To minimize historical information loss while keeping reasoning logic intact, we maintain all previous actions $\{a_i\}$ in textual form but partition the observations into a ``stale'' set $\mathcal{O}_{stale} = \{o_i\}_{i=1}^{t-k-1}$ and a ``fresh'' set $\mathcal{O}_{fresh} = \{o_j\}_{j=t-k}^{t-1}$, where $k$ is the number of most recent turns retained as text. For $\mathcal{O}_{stale}$, we render each $o_i$ individually into an image via a rendering function $\mathcal{F}_{\text{render}}$, and map it to visual tokens:

$$
\tilde{o}_i = \phi(E_v(\mathcal{F}_{\text{render}}(o_i)))
$$
where $E_v$ is the visual encoder and $\phi$ is a vision projection layer. The compressed history $\tilde{\mathcal{H}}_t$ thus accurately maintains the interleaved structure by substituting the lengthy text of stale observations with their compact visual counterparts:
$$
\tilde{\mathcal{H}}_t = \big( q, \, \{a_i, \tilde{o}_i\}_{i=1}^{t-k-1}, \, \{a_j, o_j\}_{j=t-k}^{t-1} \big)
$$

This strategy effectively transforms the $O(L)$ total token length of historical tool responses into an $O(N_v)$ length (where $N_v$ is the number of visual patches), ensuring $N_v \ll L$. On average, rendering reduces the token budget of each stale observation by about $50\%$. The rendering parameters are provided in the supplementary materials.

\smallskip \noindent \textbf{Discussions.} V-Fold features two primary advantages: (i) Adaptive Compression: Our context-aware sliding window selectively renders stale observations~($\mathcal{O}_{stale}$). Preserving the most recent $k$ interactions as text maintains the high-precision environment necessary for local reasoning continuity. (ii) Data-Efficient Visual Adaptation: Instead of relying on extensive RL or massive datasets, our approach is highly data-efficient. By incorporating a mere 5k fully rendered trajectories into the SFT mixture, the model learns to interpret visual history and seamlessly generalizes to our adaptive strategy without complex post-hoc alignment.

%% file: tables/sota.tex
\begin{table*}[t]
    \centering
    \small 
    \setlength{\tabcolsep}{2pt}
    \renewcommand{\arraystretch}{1.1}
    \begin{tabular}{l cccccc}
        \toprule
        \textbf{Model} & \textbf{MMSearch} & \textbf{MMSearch+} & \textbf{FVQA} &  \textbf{SimpleVQA} & \textbf{LiveVQA} & \textbf{BC-VL} \\
        \midrule
        \multicolumn{7}{c}{\textbf{Direct Answer}} \\
        \midrule
        GPT-5 & 33.3 & 19.1 & 57.3 & 61.7 & 57.5 & \underline{47.2} \\
        Gemini-2.5 Flash & 30.4 & 8.1 & 47.7 & 55.5 & 51.0 & 37.1 \\
        Qwen3-VL-8B-Instruct & 15.2 & 3.2 & 28.0 & 42.9 & 41.0 & 25.1 \\
        \midrule
        \multicolumn{7}{c}{\textbf{RAG Workflow}} \\ 
        \midrule
        GPT-4o & 24.1 & - & - & 61.6 & 34.0 & 13.4 \\
        Gemini-2.5 Flash & 43.9 & - & - & 68.6 & 41.3 & 13.0 \\
        Qwen3-VL-8B-Instruct & 47.3 & - & 53.6 & 62.3 & 39.3 & 29.3 \\
        \midrule
        \multicolumn{7}{c}{\textbf{Agent Workflow}} \\ 
        \midrule
        GPT-5 & 63.7 & 17.2 & 69.0 & - & 73.3 & 46.1 \\
        Gemini-2.5 Flash & 64.0 & 19.9 & 68.0 & - & 73.0 & 44.6 \\
        Claude-4-Sonnet & 67.2 & 23.1 & 69.0 & - & 69.7 & \textbf{48.6} \\
        \midrule
        \multicolumn{7}{c}{\textbf{Agentic Search MLLM}} \\ 
        \midrule
        MMSearch-R1-7B~\cite{wu2025mmsearch} & 53.8 & - & 58.4 & 57.4 & 48.4 & - \\
        DeepMMSearch-R1-7B~\cite{narayan2025deepmmsearch} & - & - & - & 55.9 & - & - \\
        WebWatcher-32B~\cite{geng2025webwatcher} & 55.3 & - & - & 59.0 & 58.7 & 27.0 \\
        DeepEyesV2~\cite{hong2025deepeyesv2} & 63.7 & - & 60.6 & 59.4 & - & - \\
        Qwen3-VL-8B-Instruct~\cite{bai2025qwen3} & 58.5 & 10.4 & 64.0 & 65.8 & 60.5 & 30.8 \\
        Skywork-R1V4-30B-A3B~\cite{zhang2025skywork} & 66.1 & - & 67.2 & - & - & 38.4 \\
        SenseNova-MARS-8B~\cite{SenseNova-MARS} & 67.8 & - & 67.1 & \textbf{70.2} & 56.2 & - \\
        MM-DeepResearch-8B~\cite{yao2026mmdeepresearch} & 67.8 & - & 69.2 & 65.9 & 65.0 & 37.9 \\
        Vision-DeepResearch-8B~\cite{huang2026vision} & 69.6 & 20.4 & 64.7 & - & 76.7 & 42.6 \\
        \rowcolor{green!10} POINTS-Seeker-8B & \underline{70.8} & \textbf{25.2} & \textbf{71.2} & 68.8 & \underline{77.7} & 44.4 \\
        \rowcolor{green!10} Qwen3-VL-8B-Instruct + Our method & \textbf{71.3} & \underline{24.3} & \underline{71.1} & \underline{69.2} & \textbf{80.5} & 43.9 \\
        \bottomrule
    \end{tabular}
    \caption{\textbf{Comparison with state-of-the-art models.}
    \label{tab:sota}}
    \vspace{-1.5em}
\end{table*}

%% file: sections/4_experiments.tex
\section{Experiments}
\label{sec:exp}

\subsection{Experimental Setup}
\label{sec:exp_setup}

\noindent \textbf{Training datasets.} For the training dataset, the text reasoning data is sourced from AM-DeepSeek-R1-Distilled-1.4M~\cite{zhao20251}, while the text agentic data is obtained from Toucan-1.5M~\cite{xu2025toucan}. The multi-modal reasoning datasets are derived from Reason-RFT~\cite{tan2025reason} and VisualWebInstruct~\cite{jia2025visualwebinstruct}. The dataset used for SFT is drawn from the training set of  LiveVQA and FVQA, supplemented with a portion of general-domain data from Honey-Data~\cite{zhang2025bee}, whereas the RL dataset consists of LiveVQA and our self-constructed VQA samples. Detailed information about these training datasets is provided in the supplementary materials.

\smallskip \noindent \textbf{Evaluation benchmarks.} We evaluate on six benchmarks: MMSearch~\cite{jiang2025mmsearch}, MMSearch-Plus~\cite{tao2025mmsearch}, FVQA~\cite{wu2025mmsearch}, SimpleVQA~\cite{cheng2025simplevqa}, LiveVQA~\cite{fu2025seeking}, and BrowseComp-VL (BC-VL)~\cite{geng2025webwatcher}. Following prior protocols~\cite{wu2025mmsearch,huang2026vision}, we filter MMSearch for image-containing QA pairs, MMSearch-Plus for single-image samples, and SimpleVQA for English instances. The remaining benchmarks use full datasets.

\input{tables/ablation_training_strategy}

\input{tables/effiency}
\input{tables/ablation_compression}

\smallskip \noindent \textbf{Tools.} Our agent utilizes three tools: web search, image search, and visit. (i) For web search, we use the Google Custom Search API or Serper API to retrieve the titles, snippets, and URLs of the top-10 relevant webpages. (ii) For image search, we use the Serper API to obtain the top-5 webpage URLs, parse them with Crawl4AI~\cite{crawl4ai2024}, and summarize goal-relevant content using Qwen3-235B-A22B~\cite{yang2025qwen3}. (iii) The visit tool follows a similar pipeline, extracting webpage content with Crawl4AI and generating a goal-focused summary with Qwen3-235B-A22B.

\smallskip \noindent \textbf{Implementation details.} POINTS-Seeker initializes its ViT and LLM with Qwen2-VL-ViT~\cite{wang2024qwen2} and Qwen3-8B-Base~\cite{yang2025qwen3}. For both Agentic Seeding and SFT, the global batch size is set to 64, and the learning rate is \(5 \times 10^{-5}\). Batch samples are flattened and concatenated into sequences up to 32k tokens. For RL training, we use a global batch size of 64 and perform 8 rollouts per sample. The learning rate is set at \(1 \times 10^{-6}\). Additionally, the $\lambda_{\text{acc}}$, $\lambda_{\text{format}}$, and $\lambda_{\text{tool}}$ are set to 0.7, 0.2, and 0.1, respectively. We empirically set $\tau$ to 8k and $k$ to 2. Further details are provided in the supplementary materials.

\subsection{Comparison with State-of-the-art}

We evaluate POINTS-Seeker against mainstream models under four settings: (i) \textbf{Direct answer}, where the model answers without external search; (ii) \textbf{RAG workflow}, where relevant information is first retrieved and then used for answer generation; (iii) \textbf{Agent workflow}, which follows a ReAct-style agent paradigm~\cite{huang2026vision}; and (iv) \textbf{Agentic search MLLM}, where the system prompt specifies the task and available tools, and the model autonomously decides when and which tools to invoke before producing the final answer. As shown in Table~\ref{tab:sota}, we can draw the following conclusions. (i) The agent workflow paradigm significantly outperforms the direct answer setting, indicating the tool invocation is essential for achieving strong performance on these benchmarks. (ii) Our model demonstrates superior performance across these benchmarks, substantially outperforming several existing open-source agentic search MLLMs. Moverover, it either surpasses or achieves performance comparable to proprietary agent workflow models. (iii) We further evaluate the generality of our approach on Qwen3-VL-8B-Instruct by applying V-Fold and the training pipeline without the agentic seeding stage. The consistent improvements highlight the robustness of our method across different backbone models.

\input{tables/general_bench}

\subsection{Ablations and Analysis}

\noindent \textbf{Ablation on our core design.} To validate the effectiveness of our training strategies and V-Fold, we conduct ablation experiments in Table~\ref{tab:ablation_training}, yielding several key observations. (i) Agentic seeding provides consistent performance gains across all benchmarks, demonstrating that injecting agentic-related data during the model's formative stages is essential for long-horizon interaction. (ii) SFT on high-quality trajectory data substantially improves basic agentic abilities, enabling more accurate tool invocation and strong performance on relatively simple benchmarks such as SimpleVQA and FVQA. (iii) RL further reduces repetitive tool calls and improves output-format reliability over the SFT-only baseline, with case studies in the supplementary material.  (iv) V-Fold provides additional improvements, especially on challenging benchmarks such as MMSearch-Plus and BrowseComp-VL, suggesting its effectiveness in handling complex long-horizon problems.

\begin{figure}[t]
  \centering
  \includegraphics[width=.48\textwidth]{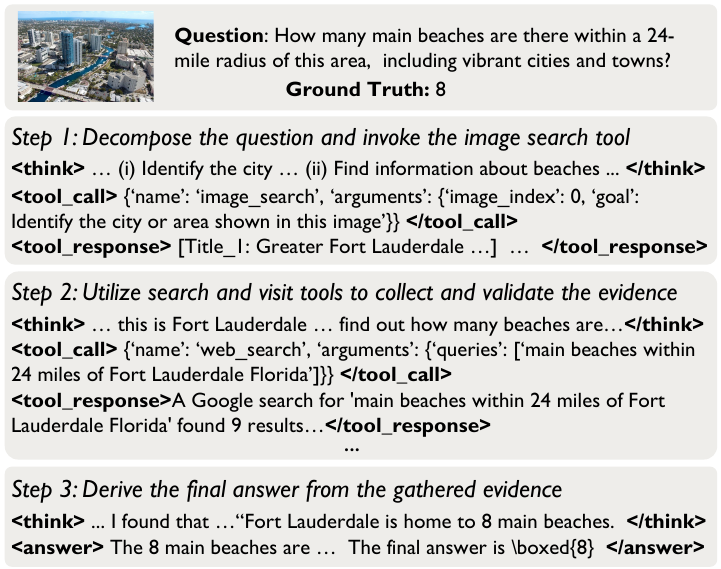} \\
  \caption{\textbf{Visualization of POINTS-Seeker.} Additional qualitative results are in the supplementary materials. }
 \label{fig:qualitative}
 \vspace{-12pt}
\end{figure}

\smallskip \noindent \textbf{In-depth analysis of V-Fold.} We further analyze how V-Fold alleviates bottlenecks in long-horizon agentic search. As shown in Table~\ref{tab:compress_ratio}, V-Fold achieves a dual optimization of efficiency and reasoning fidelity: (i) On BC-VL and MMSearch-Plus, it reduces the average response length by 22.8\% and 25.1\%, respectively, mitigating token overflow.  (ii) V-Fold's history-aware rendering preserves raw evidence from ``stale'' observations in a visual format. By folding extensive textual history into compact visual tokens, it enables the model to more effectively utilize historical cues while bypassing the performance degradation.

\smallskip \noindent \textbf{V-Fold under long-horizon settings.} To evaluate V-Fold in harder long-horizon scenarios, we use Claude-Sonnet-4-5 to construct hard subsets from each benchmark, selecting VQA instances requiring over ten interaction rounds. We compare V-Fold with three baselines: \textit{Discard History}, which keeps only the initial question and recent interactions after exceeding a threshold; \textit{Summarization}, which compresses history with Qwen3-235B-A22B; \textit{Image Render}, which converts all past observations into images. As shown in Table~\ref{tab:ablation_compression}, V-Fold outperforms these alternatives. Notably, it outperforms Image Render, highlighting the advantages of our adaptive mechanism and sliding-window strategy. This shows that selectively maintaining the recent textual context yields a better balance between compression and information retention.

\vspace{2pt}
\noindent \textbf{Impact on general visual capabilities.} 
A common concern in injecting specialized agentic skills into LMMs is the potential degradation of their basic perceptual abilities. To examine this, we evaluate POINTS-Seeker on general VQA benchmarks and compare it with POINTS-Baseline, which removes the Agentic Seeding stage and all agentic search trajectories during SFT, using only general multimodal data. As shown in Table~\ref{tab:opencompass}, POINTS-Seeker outperforms the baseline by an average of 2.3 points. This suggests that internalizing autonomous inquiry capabilities in early training does not harm general visual understanding; instead, agentic search and reasoning skills can improve performance on standard non-agentic visual tasks.

\input{tables/hyper_parameter}

\vspace{2pt}
\noindent \textbf{Ablation on hyperparameters of V-Fold.} As shown in Table~\ref{tab:hyperparameters}, we ablate the hyperparameters of V-Fold on MMSearch-Plus. The results show that an overly large $\tau$ leads to a performance drop, as it triggers compression less frequently. We also find that setting $k$ too small degrades performance, highlighting the importance of preserving recent textual context. We finally set $\tau$ to 8k and $k$ to 2.

%% file: tables/ablation_training_strategy.tex
\begin{table*}[t]
\centering
\small
\setlength{\tabcolsep}{1.5mm}{
  \begin{tabular}{cccccccccc}
    \toprule
    \textbf{Seeding} & \textbf{SFT} & \textbf{RL} & \textbf{V-Fold} & \textbf{MMSearch} & \textbf{MMSearch+} & \textbf{FVQA} & \textbf{SimpleVQA} & \textbf{LiveVQA} & \textbf{BC-VL} \\
    \midrule
    \multicolumn{10}{c}{\textbf{POINTS-Seeker}} \\ 
    \midrule
     \ding{56} & \ding{52} & \ding{56} & \ding{56} & 63.2 & 18.0 & 67.9 & 65.1 & 70.8 & 32.3 \\
     \ding{52} & \ding{52} & \ding{56} & \ding{56} & 66.1 & 23.0 & 68.9 & 66.4 & 76.8 & 39.8 \\
     \ding{52} & \ding{52} & \ding{52} & \ding{56} & 69.6 & 23.4 & 71.2 & 68.0 & 77.3 & 42.6 \\
     \ding{52} & \ding{52} & \ding{52} & \ding{52} & \textbf{70.8} & \textbf{25.2} & \textbf{71.2} & \textbf{68.8} & \textbf{77.7} & \textbf{44.4} \\
     \midrule
     \multicolumn{10}{c}{\textbf{Qwen3-VL-8B-Instruct}} \\ 
     \midrule
     - & \ding{56} & \ding{56} & \ding{56} & 58.5 & 10.4 & 64.0 & 65.8 & 60.5 & 30.8 \\
     - & \ding{52} & \ding{56} & \ding{56} & 69.0 & 23.0 & 68.9 & 67.6 & 76.9 & 42.4 \\
     - & \ding{52} & \ding{52} & \ding{56} & 70.2 & 23.0 & 70.7 & 68.5 & 80.2 & 42.6 \\
     - & \ding{52} & \ding{52} & \ding{52} & \textbf{71.3} & \textbf{24.3} & \textbf{71.1} & \textbf{69.2} & \textbf{80.5} & \textbf{43.9} \\
    \bottomrule
  \end{tabular}
  \caption{\textbf{Ablation study on our core design.}}
  \vspace{-8pt}
  \label{tab:ablation_training}
}
\end{table*}

%% file: tables/effiency.tex
\begin{table}
    \centering
    \small
    \begin{tabular}{lcll}
        \toprule
        \textbf{Method} & \textbf{Dataset} & \textbf{Acc} & \textbf{Resp. Len.}  \\
        \midrule
         w/o V-Fold & BC-VL & 42.6 & 9530 \\
         V-Fold & BC-VL & 44.4\textsubscript{\scalebox{0.8}{\textcolor{DeepGreen}{(1.8$\uparrow$)}}} & 7353\textsubscript{\scalebox{0.8}{\textcolor{DeepGreen}{(22.8\%$\downarrow$)}}} \\
         \midrule 
         w/o V-Fold & MMSearch+ & 23.4 & 9178 \\
         V-Fold & MMSearch+ & 25.2\textsubscript{\scalebox{0.8}{\textcolor{DeepGreen}{(1.8$\uparrow$)}}} & 6871\textsubscript{\scalebox{0.8}{\textcolor{DeepGreen}{(25.1\%$\downarrow$)}}} \\
        \bottomrule
  \end{tabular}
    \caption{\textbf{Impact of V-Fold on length and accuracy.}}
    \label{tab:compress_ratio}
\end{table}

%% file: tables/ablation_compression.tex
\begin{table}[t]
\centering
\small
\setlength{\tabcolsep}{0.8mm}{
  \begin{tabular}{lcccc}
    \toprule
    \textbf{Method} & \textbf{FVQA-H} & \textbf{LiveVQA-H} & \textbf{BC-VL-H} \\
    \midrule
     No Compression & 48.0 & 48.8 & 21.3 \\
     Discard History & 52.0 & 50.2 & 21.3 \\
     Summarization & 50.0 & 48.8 & 22.7 \\
     Image Rendering & 48.0 & 44.4 & 17.3 \\
     Our V-Fold & 54.0 & 50.7 & 26.7 \\
    \bottomrule
  \end{tabular}
  \caption{\textbf{Comparison under long-horizon settings.}}
  \vspace{-12pt}
  \label{tab:ablation_compression}
}
\end{table}

%% file: tables/general_bench.tex
\begin{table*}[t]
\centering 
\small
\setlength{\tabcolsep}{3pt} 
\renewcommand{\arraystretch}{1.2} 
\begin{tabular}{l c c c c c c c c c}
\toprule
\textbf{Method} & \textbf{MMBench} & \textbf{MMStar} & \textbf{MMMU} & \textbf{MathVista} & \textbf{Hallu.} & \textbf{AI2D} & \textbf{OCRBench} & \textbf{MMVet} & \textbf{Avg.} \\

\midrule
POINTS-Baseline & 74.9 & 55.5 & 52.4 & 75.7 & 53.2 & 81.6 & 85.1 & 67.8 & 68.3 \\
POINTS-Seeker & 79.1 & 59.7 & 52.6 & 73.7 & 54.2 & 84.8 & 86.7 & 73.9 & 70.6 \\
\bottomrule
\end{tabular}
\caption{\textbf{Performance on general VQA benchmarks.}}
\label{tab:opencompass}
\vspace{-1em}
\end{table*}

%% file: tables/hyper_parameter.tex

\begin{table}[t]
\centering 
\vspace{-0.2cm}
\small
\setlength{\tabcolsep}{5pt} 
\renewcommand{\arraystretch}{1.2} 
\begin{tabular}{c c c c c c c}
\toprule
\textbf{$\tau$} & 4k & 8k & 12k & 8k & 8k & 8k \\
\textbf{$k$} & 2 & 2 & 2 & 0 & 2 & 4  \\
Acc. & 25.2 & 25.2 & 23.4 & 23.4 & 25.2 & 25.2 \\
\bottomrule
\end{tabular}
\caption{\textbf{Ablation study on hyperparameters.}}
\label{tab:hyperparameters}
\vspace{-16pt}
\end{table}

%% file: sections/5_conclusion.tex
\section{Conclusion}
\label{sec:conclusion}


In this paper, we present \textbf{POINTS-Seeker-8B}, a multimodal search agent built from a vanilla, non-agentic vision-language model. With a dedicated dataset and training recipe, the model acquires essential agentic search capabilities. To handle extended interaction histories, we further propose \textbf{V-Fold}, a history-aware memory folding mechanism that preserves stale but critical evidence in the visual space and alleviates attention dilution. Experiments on six benchmarks show that POINTS-Seeker-8B consistently achieves state-of-the-art performance among models of the same size, demonstrating the effectiveness of our framework. We hope our work provides a practical foundation for developing native agentic LMMs capable of active exploration and reliable reasoning.

%% file: sections/limitation.tex
\section*{Limitations}


Despite its effectiveness, our approach has several limitations. First, the current context window is constrained to 32k tokens with a maximum of 10 interaction turns, which may limit performance in extremely long-horizon reasoning tasks. Second, our toolset is currently restricted to three core search tools; incorporating more specialized APIs, such as Google Scholar or Google Maps, could further broaden the model's utility. Finally, due to search API quota constraints, the scale of our RL training was limited. Future work will focus on scaling the reinforcement learning process and expanding the tool-use diversity to further enhance the model's autonomous search capabilities.

%% file: sections/X_supp.tex
\appendix

\section{Details about the Training Stages}

\subsection{Alignment \& Pre-training}

\noindent \textbf{Alignment stage.} During this initial phase, we keep the parameters of both the vision encoder and the LLM frozen, updating only the projection layer ({\em i.e.}, the MLP adapter). The training corpus for this stage consists of approximately 3M samples curated from Captioning and OCR datasets, primarily sourced from Laion-5B~\cite{schuhmann2022laion} and OpenVid~\cite{nan2024openvid}. To enhance the quality, the raw metadata from these datasets are re-annotated by state-of-the-art vision-language models~({\em e.g.,} InternVL3~\cite{zhu2025internvl3}).

\vspace{2pt}
\noindent \textbf{Pre-training stage.} Following alignment, we perform large-scale pre-training to enhance the model's multimodal understanding and document-level reasoning. We download all PDFs from the CC-MAIN-2021-31-PDF-UNTRUNCATED dataset\footnote{\url{https://digitalcorpora.org/corpora/file-corpora/cc-main-2021-31-pdf-untruncated/}} and filter for documents in Chinese and English. To construct an image-text dataset for training, we develop an automated pipeline comprising layout detection and text extraction. For layout detection, we employ PP-DocLayout~\cite{cui2025paddleocr} to extract all images from a page. Subsequently, we utilize a document OCR model~({\em i.e.}, POINTS-Reader~\cite{liu2025points}) to extract all text from the page. Finally, the images and text associated with each document are concatenated, with images positioned at the beginning. The resulting pre-training dataset contains approximately 400 billion tokens. In contrast to the alignment stage, both the vision projector layer and the LLM are jointly optimized during this phase.

\subsection{Agentic Seeding}

As detailed in Section~\ref{subsec:agentic_seeding} of the main paper, this stage focuses on instilling the model with complex reasoning and planning capabilities. We curate a diverse training suite comprising AM-DeepSeek-R1-Distilled-1.4M, Toucan-1.5M, Reason-RFT, and VisualWebInstruct. For multi-modal datasets, we utilize only the images and questions, and then distill the reasoning chain using \texttt{Qwen3-VL-235B-A22B-Thinking}, filtering based on the original answer.

\begin{itemize}[label={\bf $\bullet$}, leftmargin=*, topsep=0.1ex, itemsep=-0.1ex]
\item \textbf{AM-DeepSeek-R1-Distilled-1.4M}: This dataset provides approximately 1.4 million high-quality, challenging reasoning problems designed to provoke deep chain-of-thought processing.
\item \textbf{Toucan-1.5M}: A tool-agent dataset featuring 1.5 million trajectories. These are synthesized from 495 real-world Model Context Protocols (MCP), spanning over 2,000 unique tools to bolster the model’s agentic proficiency.
\item \textbf{Reasoning-RFT}: A multimodal reasoning dataset containing over 10,000 samples. Its construction follows a rigorous three-stage pipeline: automated generation, filtering, and human verification to ensure logical groundedness.
\item \textbf{VisualWebInstruct}: Consisting of approximately 900,000 question-answer (QA) pairs, this dataset specifically addresses the scarcity of high-quality, diverse training data for complex multimodal reasoning tasks.
\end{itemize}

\subsection{Supervised Fine-tuning}

As introduced in Section~\ref{subsec:sft}, the SFT stage aims to align the model's intrinsic agentic capabilities with the precise execution of search-related tools. Following the methodology described in Section~\ref{subsec:data}, we construct a multimodal VQA dataset comprising 5,339 high-quality trajectories. To facilitate the V-Fold mechanism, these trajectories are transformed into a dual-format representation: for each VQA sample, we provide both a text-only trajectory and a rendered visual trajectory where all intermediate tool observations are rendered into images to enable context-efficient visual processing. This data is further augmented with some QA trajectories from the InfoSeek dataset~\cite{xia2025open} and 15k OCR-specific samples to ensure high-fidelity text recognition. Finally, to maintain broad-spectrum instruction-following capabilities and prevent catastrophic forgetting, we incorporate Honey-Data-1M~\cite{wei2025open} as a general multimodal foundation. For samples in Honey-Data containing existing reasoning traces ({\em i.e.}, \texttt{<think>} tags), we employ Qwen3-VL-235B-A22B-Thinking to re-annotate the chain-of-thought content, followed by a rigorous filtering process to exclude any samples where the re-annotated answer deviates from the original response provided in the dataset. The training is conducted using our in-house framework, which is analogous to Megatron~\cite{shoeybi2019megatron}.

\subsection{Reinforcement Learning}

Building upon the SFT stage, we perform reinforcement learning to further refine the model's decision-making and tool-use precision. Our RL framework is implemented using the verl~\cite{sheng2024hybridflow}, integrated with the SGLang~\cite{zheng2024sglang} inference engine to accelerate the rollout process. The training is conducted for 60 steps, with a maximum trajectory depth of 10 turns per episode. The training corpus for this phase is composed of our self-constructed reasoning tasks and an adapted version of the LiveVQA dataset. To encourage more robust and grounded reasoning, we reformulate a significant portion of the originally multiple-choice LiveVQA samples into open-ended questions. Crucially, we bypass V-Fold during RL training. Empirical results, as detailed in Table~\ref{tab:v_fold_rl}, demonstrate that employing V-Fold during the RL phase yields performance comparable to training without compression. Consequently, we opt for the more generalized uncompressed approach for RL training to maintain simplicity and robustness.

\input{supp_tables/V-Fold_RL}

\section{More Implementation Details}

\subsection{Model Architecture}

Our model architecture is initialized with Qwen3-8B-Base~\cite{yang2025qwen3} and Qwen2-VL-ViT~\cite{wang2024qwen2}. For the positional encoding, the model applies 1D RoPE~\cite{su2024roformer} for visual tokens within the LLM backbone and 2D RoPE within the ViT image encoder. Furthermore, the intermediate projector incorporates a pixel-shuffle operation, effectively reducing the visual sequence length by a factor of 4.

\subsection{Rendering Parameters}

For image rendering, we adopt the parameter configurations from Glyph~\cite{cheng2025glyph}. The detailed specifications of these rendering parameters are summarized in Table~\ref{tab:render_param}.

\input{supp_tables/render_config}

\subsection{Computation Resources}
We utilize 128×H20 GPUs for both the agentic seeding stage and the SFT stage. The agentic seeding stage runs for approximately 18 hours, while the SFT stage takes around 7 hours. For the subsequent RL stage, we employ 64× H20 GPUs, with a total execution time of around 24 hours. For benchmarks with fewer than 300 samples, we report the average results on 3 trials.

\subsection{Prompts}

Our system prompt and three tools schema are as follows.
\input{supp_tables/system_prompt}
\input{supp_tables/web_search_schema}
\input{supp_tables/visit_schema}
\input{supp_tables/image_search_schema}
\noindent Our QA2VQA prompt is shown below.
\input{supp_tables/qa2vqa_prompt}
\noindent Our answer filtering prompt is as follows.
\input{supp_tables/answer_filtering}

\begin{figure*}[t]
  \centering
  \includegraphics[width=\textwidth]{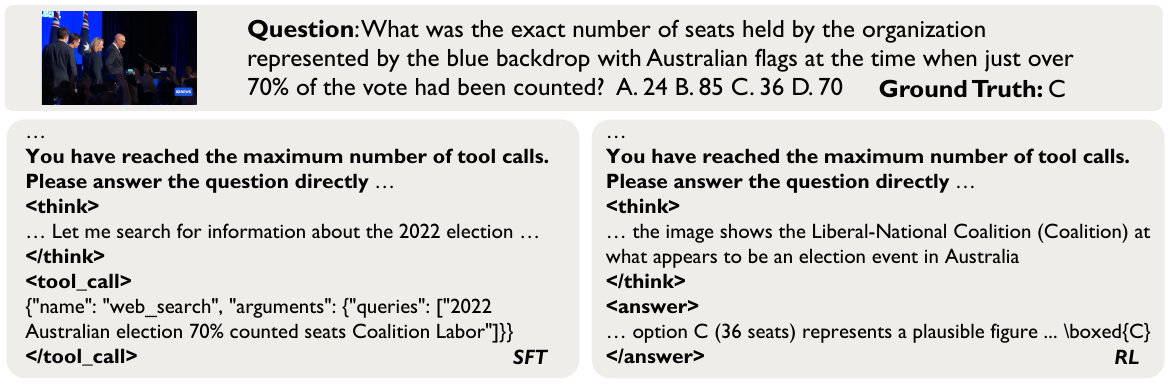} \\
  \caption{\textbf{Case Study 1: SFT vs. RL.} Details are simplified for visualization.}
 \label{fig:case_study1}
 \vspace{-6pt}
\end{figure*}

\begin{figure*}[t]
  \centering
  \includegraphics[width=\textwidth]{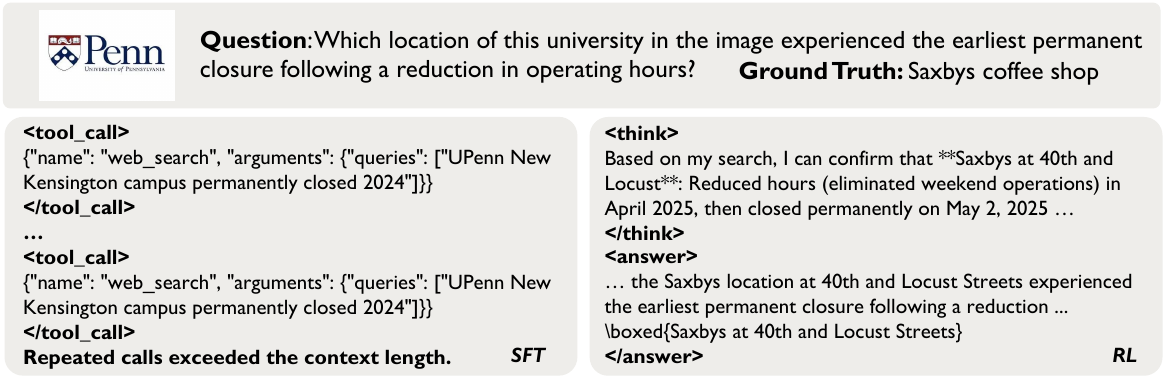} \\
  \caption{\textbf{Case Study 2: SFT vs. RL.} Details are simplified for visualization.}
 \label{fig:case_study2}
 \vspace{-6pt}
\end{figure*}

\section{More Qualitative Results}

\subsection{Case Study}
Figure~\ref{fig:case_study1} and Figure~\ref{fig:case_study2} compare the model's behavior before and after RL training. We observe that while the SFT model occasionally suffers from weak instruction-following and redundant tool-calls, RL training effectively rectifies these issues. The RL-refined model exhibits significantly better logical rigor and systematic planning, ensuring efficient, non-repetitive tool usage in complex search tasks.

\begin{figure*}[t]
    \centering
    \begin{subfigure}[t]{0.43\textwidth}
        \centering
        \includegraphics[
            width=\linewidth,
            height=5cm,
            keepaspectratio
        ]{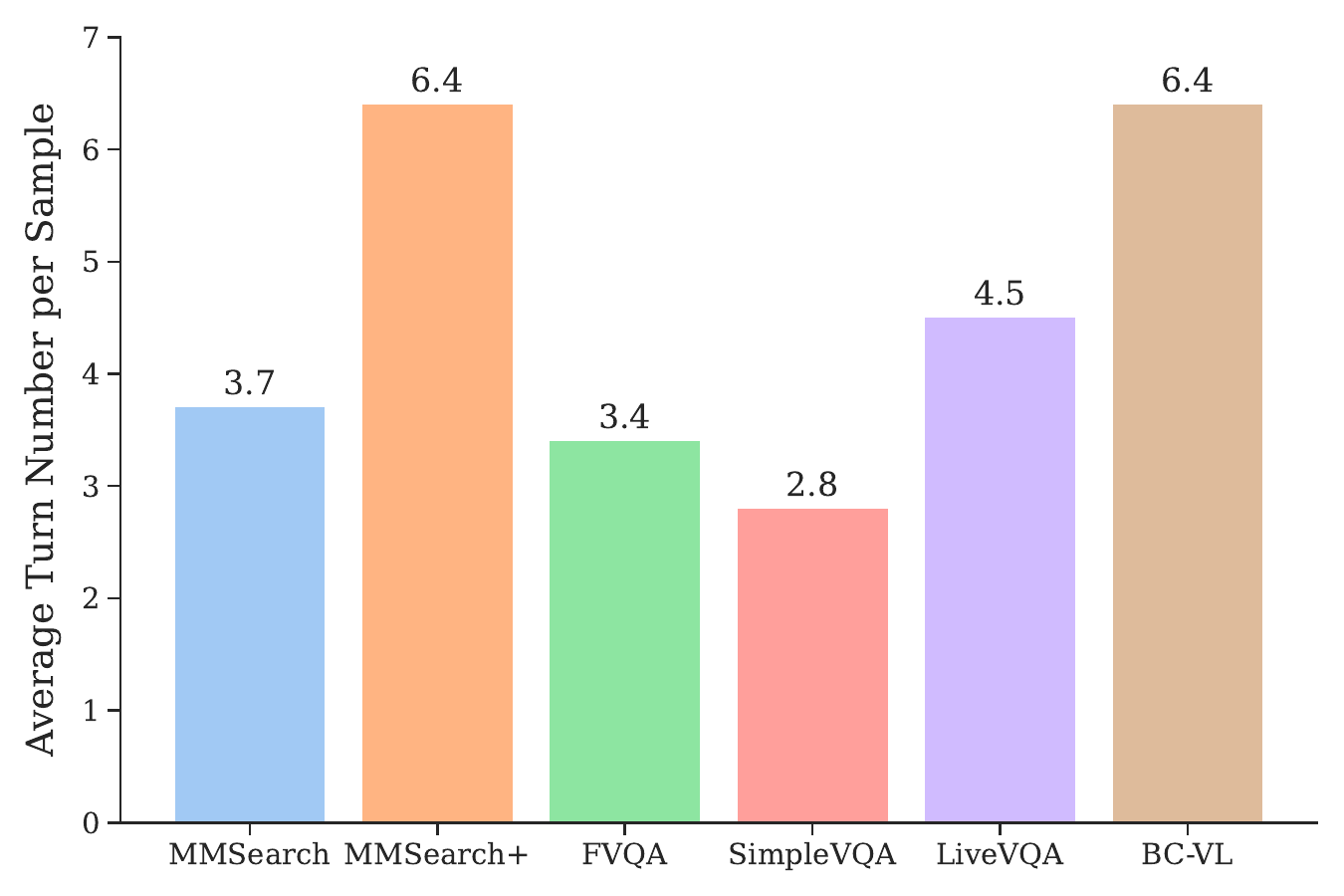}
        \caption{Average number of tool interaction turns.}
        \label{fig:tool_call_distribution}
    \end{subfigure}
    \hfill
    \begin{subfigure}[t]{0.55\textwidth}
        \centering
        \includegraphics[
            width=\linewidth,
            height=5cm,
            keepaspectratio
        ]{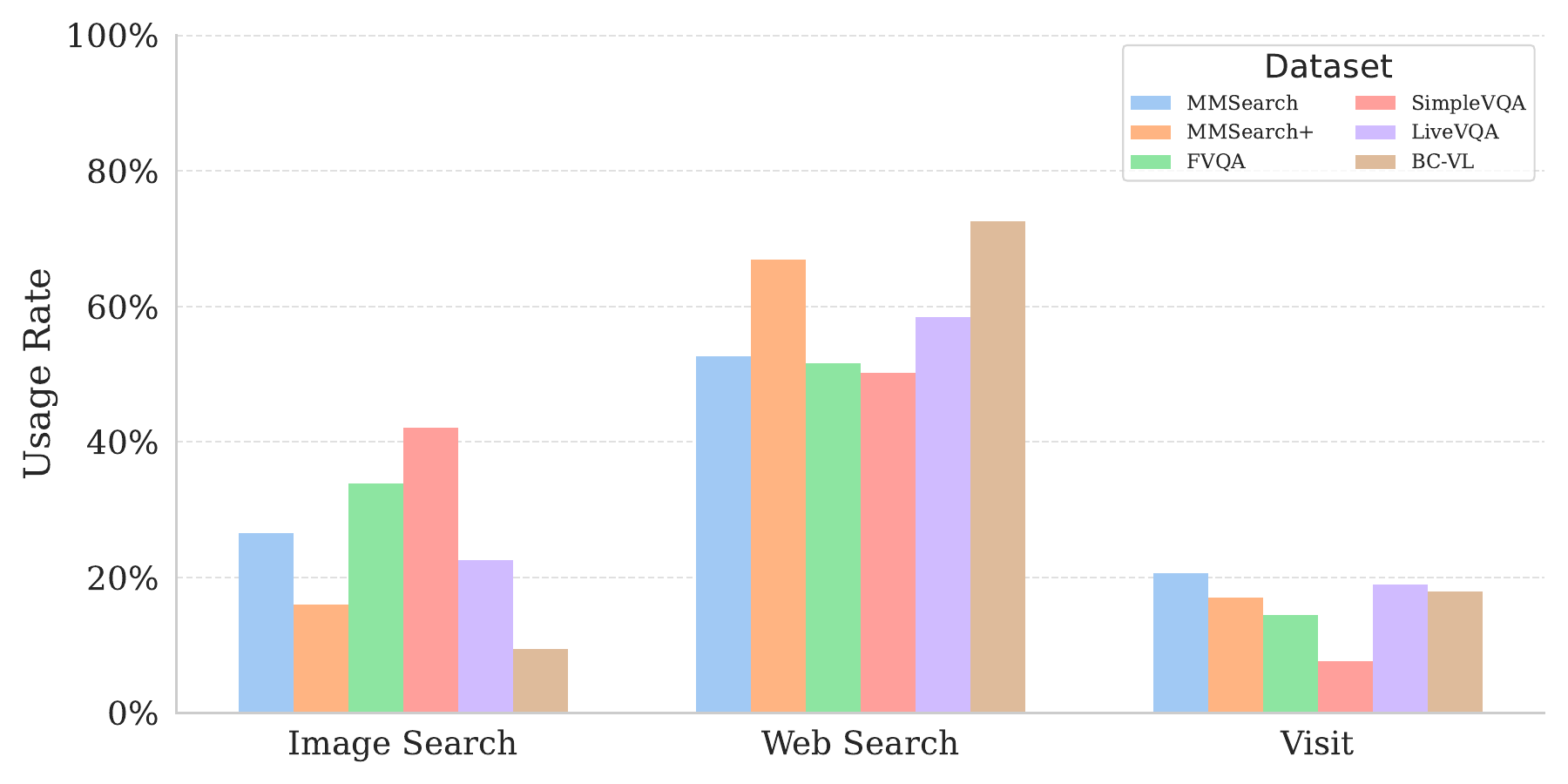}
        \caption{Percentage distribution of tool types across various benchmarks.}
        \label{fig:tool_type_distribution}
    \end{subfigure}
    \caption{\textbf{Tool usage analysis for POINTS-Seeker across six diverse evaluation benchmarks.}}
    \vspace{-2pt}
    \label{fig:tool_analysis}
\end{figure*}

\subsection{Analysis of tool usage.} As shown in Figure~\ref{fig:tool_analysis}, we analyze the tool interaction patterns of POINTS-Seeker. We can draw the following observations: (i) POINTS-Seeker can flexibly adjust the number of interaction turns according to the difficulty of different benchmarks. For relatively simpler benchmarks such as SimpleVQA, the average number of interaction turns is 2.8. In contrast, for more challenging benchmarks such as BrowseComp-VL, the average number increases substantially to 6.4. (ii) Across all benchmarks, the model consistently exhibits a preference for using the web search tool, whereas the image search and visit tools are employed less frequently. This observation is intuitive: image search can be invoked at most once per sample, and the visit tool can only be triggered after a valid URL is provided by another tool. (iii) Tool usage varies by benchmark characteristics. For instance, in FVQA, the model relies more heavily on image search to identify rare entities, whereas in BrowseComp-VL, a higher proportion of queries are resolved through textual search or direct image interpretation.

\subsection{Qualitative Results}
\input{supp_tables/qualitative_2}
\input{supp_tables/qualitative_1}

\section{Ethical Considerations}
\label{sec:ethical}

\subsection{Potential Risks}
\label{subsec:risks}

POINTS-Seeker-8B may inherit risks from both web retrieval and generative modeling. Retrieved content can be noisy, outdated, biased, or maliciously crafted, potentially causing incorrect answers or prompt-injection vulnerabilities. The agent may also encounter sensitive or copyrighted information, raising privacy and misuse concerns, and may still hallucinate when the evidence is insufficient. Deployment should therefore incorporate source verification, privacy filtering, prompt-injection defense, tool-use sandboxing, and human oversight in high-stakes scenarios.

\subsection{License for Artifacts}
\label{subsec:license}
We will release the code, model checkpoints, training/evaluation scripts, and curated data under appropriate open-source licenses. The code will be released under the Apache-2.0 license. The released model checkpoints will follow the license of the corresponding backbone model and will be provided for research use. The curated datasets and trajectories will be released under a research-friendly license, subject to the licenses and terms of the original data sources. For any third-party resources used in this work, including backbone models, benchmarks, and external tools, we comply with their respective licenses and usage terms.

\subsection{Data Safety and Anonymization}
\label{subsec:safety}

During data curation, we did not intentionally collect private or sensitive personal information. For data derived from web sources, we avoid redistributing raw content when privacy or licensing concerns are unclear, and release only permitted artifacts or derived metadata when applicable.

\subsection{Artifact Documentation}
The data mainly covers knowledge-intensive multimodal search scenarios such as entities, landmarks, products, web pages, charts, documents, and multi-hop VQA. The language coverage is primarily English, with possible multilingual content inherited from public web sources and benchmarks. The artifacts are intended for research on multimodal search agents and long-horizon reasoning. We do not intentionally collect or annotate demographic attributes.

\subsection{Use of AI Assistants.}
We used AI assistants to help with language polishing and improving the clarity of the manuscript. The authors are fully responsible for all technical content, experimental results, analyses, and claims in the paper.

%% file: supp_tables/V-Fold_RL.tex
\begin{table}[t]
\centering
\scriptsize
\setlength{\tabcolsep}{0.3mm}{
  \begin{tabular}{ccccccc}
    \toprule
    \textbf{Method} & \textbf{MMSearch} & \textbf{MMSearch+} & \textbf{FVQA} & \textbf{SimpleVQA} & \textbf{LiveVQA} & \textbf{BC-VL} \\
    \midrule
     V-Fold RL & 70.8 & 24.3 & 70.6 & 68.0 & 78.5 & 42.1 \\
     Our RL & 70.8 & 25.2 & 71.2 & 68.8 & 77.7 & 44.4 \\
    \bottomrule
  \end{tabular}
  \caption{\textbf{Ablation study on RL training strategy.}}
  \label{tab:v_fold_rl}
  \vspace{-8pt}
}
\end{table}

%% file: supp_tables/render_config.tex

\begin{table}[htbp]
\centering
\small
\setlength{\tabcolsep}{8mm}{
\begin{tabularx}{.48\textwidth}{l X}
\toprule
\textbf{Parameter} & \textbf{Value} \\
\midrule
\texttt{page-size}             & $595 \times 842$ \\
\texttt{dpi}                   & 72 \\
\texttt{margin-x}              & 10 \\
\texttt{margin-y}              & 10 \\
\texttt{font-path}             & \texttt{Verdana.ttf} \\
\texttt{font-size}             & 9 \\
\texttt{line-height}           & 10 \\
\texttt{font-color}            & \texttt{\#000000} \\
\texttt{alignment}             & \texttt{LEFT} \\
\texttt{horizontal-scale}      & 1.0 \\
\texttt{first-line-indent}     & 0 \\
\texttt{left-indent}           & 0 \\
\texttt{right-indent}          & 0 \\
\texttt{space-before}          & 0 \\
\texttt{space-after}           & 0 \\
\texttt{border-width}          & 0 \\
\texttt{border-padding}        & 0 \\
\texttt{page-bg-color}         & \texttt{\#FFFFFF} \\
\texttt{para-bg-color}         & \texttt{\#FFFFFF} \\
\texttt{auto-crop-width}       & \texttt{true} \\
\texttt{auto-crop-last-page}   & \texttt{true} \\
\bottomrule
\end{tabularx}
\caption{Parameter configuration for image rendering.}
\label{tab:render_param}}
\end{table}

%% file: supp_tables/system_prompt.tex
\begin{jsonbox}
You are a specialized multimodal agent. Your purpose is to solve visual question answering tasks by thinking step-by-step and using tools.

# Tools
You may call one or more functions to assist with the user query.
You are provided with function signatures within <tools></tools> XML tags:
<tools>
[Detailed Tool Schemas (web_search, visit, image_search) will be dynamically injected here. Refer to the specific functional definitions for parameter constraints.]
</tools>

For each function call, return a json object with function name and arguments within <tool_call></tool_call> XML tags:
<tool_call>
{{"name": <function-name>, "arguments": <args-json-object>}}
</tool_call>

# Instruction
1. In each turn, you should start with <think> tag. In this tag, you need to conduct a step-by-step reasoning process about the image and question and evaluate whether tool use would be helpful and give the reason. If received tool results, you also need to analyze them.
2. If you think some tools are useful, call them in <tool_call> tag. **Remember: You can only make one single tool call per turn.**
3. If you think no more tools are needed, you can answer in <answer> tag. You need to provide a concise summary of your reasoning process that leads to the final answer. Besides, you also need to put a simple and direct answer in \boxed{{}} for verification.

The structure of your response should be like this:
<think> thinking process satisfying the Instruction 1 </think>
(<tool_call> tool calls satisfying the Instruction 2 </tool_call> | <answer> answer satisfying the Instruction 3 </answer> )

**Search Trigger**: If you are identifying a specific entity (building, person, artwork), you **must** use the image_search or web_search tool. 
**Visit for Detail**: Use the visit tool to scrape specific webpage content if the search snippet is insufficient.
**Repeated Verification**: Before presenting a Final Answer, you will **cross-check** and **validate the information** you've gathered to confirm its accuracy and reliability.
**Question Decomposition**: Break the user query into atomic sub-questions. (e.g., 1. Identify the building. 2. Find its height. 3. Compare with the user's target).
\end{jsonbox}

%% file: supp_tables/web_search_schema.tex
\begin{jsonbox}
{
  "type": "function",
  "function": {
    "name": "web_search",
    "description": "Call this tool to interact with the web_search API. You will receive the top 10 text excerpts from Google's text search engine using text as the search query.",
    "parameters": {
      "type": "object",
      "properties": {
        "queries": {
          "type": "array",
          "items": {
            "type": "string",
            "description": "The search query."
          },
          "description": "The list of search queries."
        }
      },
      "required": ["queries"]
    }
  }
}
\end{jsonbox}

%% file: supp_tables/visit_schema.tex
\begin{jsonbox}
{
  "type": "function",
  "function": {
    "name": "visit",
    "description": "Visit webpage(s) and return the summary of the content.",
    "parameters": {
      "type": "object",
      "properties": {
        "url": {
          "type": "array",
          "items": {
            "type": "string",
            "description": "The url of the webpage to visit."
          },
          "description": "The list of search urls."
        },
        "goal": {
          "type": "string",
          "description": "The specific information goal for visiting webpage(s)."
        }
      },
      "required": ["url", "goal"]
    }
  }
}
\end{jsonbox}

%% file: supp_tables/image_search_schema.tex
\begin{jsonbox}
{
  "type": "function",
  "function": {
    "name": "image_search",
    "description": "Utilize the search engine to retrieve relevant information based on the input image and return results related to the goal.",
    "parameters": {
      "type": "object",
      "properties": {
        "image_index": {
          "type": "integer",
          "description": "The index of the image you want to search for more information."
        },
        "goal": {
          "type": "string",
          "description": "The specific information goal."
        }
      },
      "required": ["image_index", "goal"]
    }
  }
}
\end{jsonbox}

%% file: supp_tables/qa2vqa_prompt.tex
\begin{jsonbox}
You are a Multimodal Data Engineer. Your goal is to convert a QA pair into a VQA query where the image acts as a necessary visual context, NOT as a direct hint to the answer.

### STRICT RULES:
1. TARGET ENTITY != ANSWER: The entity you select to mask (the Target Entity) MUST NOT be the same as the Answer.
2. VISUAL ANCHORING: Select an entity from the QUESTION that serves as a visual anchor (e.g., a location, a related object, or the subject's environment).
3. MAINTAIN COMPLEXITY: Keep all the complex reasoning steps from the original question.
4. MASKING: Replace the selected Target Entity with a visual reference token (e.g., "this location", "this botanical garden", "the habitat shown here").

### OUTPUT FORMAT (JSON ONLY):
{
  "status": "success" | "discard",
  "selected_target_entity": "The entity chosen from the QUESTION (must not be the answer)",
  "transformed_vqa_query": "The rewritten complex question",
  "original_answer": "{{Answer}}",
  "search_query": "Search string for the selected_target_entity"
}

### INPUT:
Question: [Question]
Answer: [Answer]
\end{jsonbox}

%% file: supp_tables/answer_filtering.tex
\begin{jsonbox}
### Role 
You are a rigorous Semantic Evaluation Expert. Your task is to determine if the "Model Output" is factually and semantically consistent with the "Ground Truth."

### Evaluation Criteria
1. **Semantic Equivalence**: Respond "yes" if the core meaning, facts, and conclusions are the same, even if wording, syntax, or language differs.
2. **Detail Level**: If the Model Output is more detailed than the Ground Truth but maintains the same core fact, it is consistent ("yes").
3. **Fact Conflict**: Respond "no" if there is a contradiction, a wrong numerical value, or a different entity mentioned.
4. **Format Neutrality**: Ignore casing, punctuation, and minor spelling errors.

### Example 1
- **Question**: "What official residence was directed to be built by a colonial Governor of Dominica and whose gardens were damaged by hurricanes in 1916, 1930, and 1979? Note: The residence was built near the location shown in the image."
- **Ground Truth**: "Government House, Dominica"
- **Model Output**: "Government House"
- **Respond**: "yes"

### Example 2
- **Question**: "What NFL season ended with the Giants defeating the Chicago Bears 47:7 in the championship game shown here, which was coached by a person whose offensive coordinator was Vince Lombardi, and the game was played on an icy field due to cold weather conditions, and this Vince Lombardi later coached a team that won five total NFL Championships in seven years?"
- **Ground Truth**: "1956 New York Giants season"
- **Model Output**: "1956"
- **Respond**: "yes"

### Context
- **Question**: {question}
- **Ground Truth**: {gt_answer}
- **Model Output**: {output_answer}

### Task
Is the Model Output consistent with the Ground Truth? Respond only with "yes" or "no".
\end{jsonbox}

%% file: supp_tables/qualitative_2.tex
\begin{tcolorbox}[breakable,title=Case Trajectory in MMSearch]
\fontfamily{lmtt}\selectfont
\raggedright
\textcolor{black}{\textbf{Question:} Who is the man wearing the number 6 jersey?}\\
\textcolor{black}{\textbf{Image:}}

\begin{center}
  \includegraphics[width=0.8\linewidth]{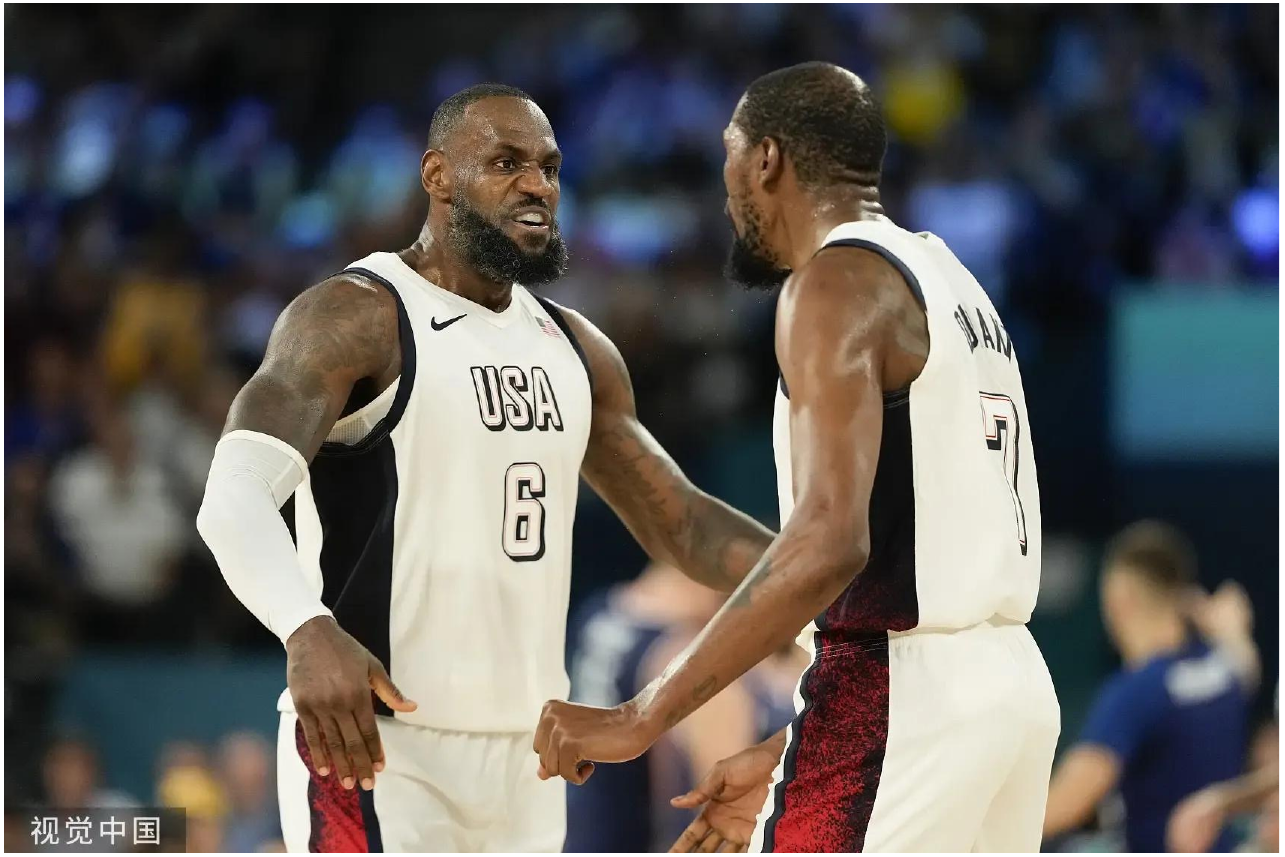}
\end{center}

\colorbox{gray!70}{%
    \parbox{\dimexpr\textwidth-2\fboxsep}{%
    \centering
        \textbf{Step 1}%
    }%
}
\textcolor{black}{\textbf{<think>} \\
I need to identify the basketball player wearing the number 6 jersey in this image. The image shows two players in USA basketball uniforms, with one wearing number 6. This appears to be from a USA team game or tournament.}\\
\textcolor{black}{To identify this specific player, I should use the image\_search tool to find out who wears number 6 for the USA basketball team.\\
\textbf{</think>}}\\
\textcolor{black}{\textbf{<tool\_call>} \\ 
\{"name": "image\_search", "arguments": \{"image\_index": 0, "goal": "Identify the basketball player wearing number 6 jersey for the USA team"\}\}
\\
\textbf{</tool\_call>}}\\
\textbf{<tool\_response>}\\
\texttt{[}Title\_1: Chansky's Notebook: Heartfelt Win - Chapelboro.com\texttt{]}(https://chapelboro .com/sports/chanskys-notebook-heart felt-win)\\
Summary\_1: 
The webpage discusses the performances of key players on Team USA, including LeBron James, Kevin Durant, Stephen Curry, and Joel Embiid, during a crucial game against Serbia. However, there is no specific mention of the player wearing the number 6 jersey for the USA team, making it impossible to identify based on the provided content.\\
\texttt{[}Title\_2: LeBron James, still leading the way for the US at age 39 ...\texttt{]}(https://www.wjhl.com/sports/us\\-world-sports/ap-lebron-james-still-\\leading-the-way-for-the-us-at-age-\\39-has-another-chance-at-an-olympic\\-gold/)\\
Summary\_2: Failed to parse the web page.\\
\texttt{[}Title\_3: U.S. Men's basketball team advances after beating Serbia ...\texttt{]}(https://www.youtube.com/watch\\?v=eisXkbYVt8c)\\
Summary\_3: 
The webpage content includes multiple references to the U.S. Men's basketball team's performance against Serbia focusing on highlights, commentary ...


...\\
\textbf{</tool\_response>}\\
\colorbox{gray!70}{%
    \parbox{\dimexpr\textwidth-2\fboxsep}{%
    \centering
        \textbf{Step 2}%
    }%
}\\

\textcolor{black}{\textbf{<think>} \\
Based on the search results, I found that the player wearing number 6 for the USA team is LeBron James. Multiple sources confirm this: 1. The first result mentions "LeBron James, Kevin Durant, and Stephen Curry" as key players for Team USA 2. The fourth result specifically states "LeBron James to be Team USA's male flag bearer for Olympic" 3. The fifth result explicitly states "the basketball player wearing the number 6 jersey for the USA team is LeBron James"\\
The image shows two players in USA uniforms - one wearing number 6 (LeBron James) and another wearing number 7 (Kevin Durant, as visible on the back of his jersey). This appears to be from a USA team game, likely during the Olympics based on the context from the search results.\\
I have enough information to provide a definitive answer.\\
\textbf{</think>}}\\
\textcolor{black}{\textbf{<answer>} \\ 
Based on the image and the search results, the man wearing the number 6 jersey is **LeBron James**. He is a legendary NBA player who has been a key member of the USA men's basketball team, serving as the team's male flag bearer for the Paris Olympics. The image shows him in a USA team uniform alongside Kevin Durant (wearing number 7), likely during an Olympic basketball game.\\
\textbackslash boxed\{LeBron James\} \textcolor{black}{\checkmark}
\\
\textbf{</answer>}}\\
\end{tcolorbox}

%% file: supp_tables/qualitative_1.tex
\begin{tcolorbox}[breakable,title=Case Trajectory in BrowseComp-VL]
\fontfamily{lmtt}\selectfont
\raggedright
\textcolor{black}{\textbf{Question:} What is the dimension of this figure, which is constructed by recursively removing the middle third from a line segment and listed by Hausdorff dimension?}\\
\textcolor{black}{\textbf{Image:}}

\begin{center}
  \includegraphics[width=0.8\linewidth]{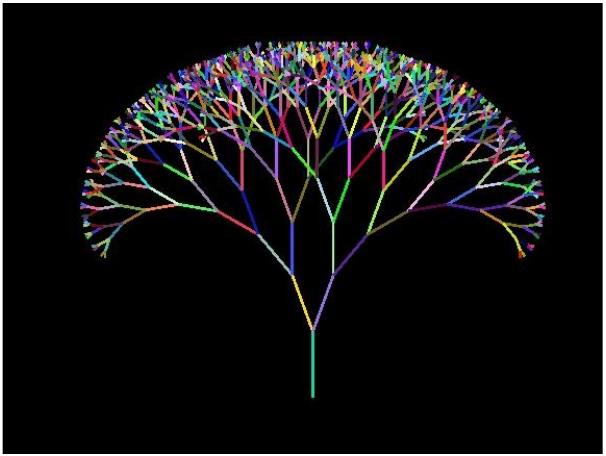}
\end{center}

\colorbox{gray!70}{%
    \parbox{\dimexpr\textwidth-2\fboxsep}{%
    \centering
        \textbf{Step 1}%
    }%
}
\textcolor{black}{\textbf{<think>} \\
Let me analyze this image. I can see a fractal-like structure that appears to be a tree or branching pattern with many colored lines. The description mentions it's constructed by "recursively removing the middle third from a line segment" and is "listed by Hausdorff dimension." This sounds like it could be related to the Cantor set or a similar fractal construction. When you recursively remove the middle third from a line segment, you're creating a Cantor set. The Hausdorff dimension of the Cantor set is a well-known mathematical concept. However, I should verify what specific fractal this is and what its Hausdorff dimension is. The image shows a tree-like structure with branching, which might be a different fractal construction. Let me search for more information about this. I need to identify what this specific figure represents and find its Hausdorff dimension.\\
\textbf{</think>}}\\
\textcolor{black}{\textbf{<tool\_call>} \\ 
\{"name": "image\_search", "arguments": \{"image\_index": 0, "goal": "Identify this fractal figure and find its Hausdorff dimension"\}\}
\\
\textbf{</tool\_call>}}\\
\textbf{<tool\_response>}\\
\begin{center}
  \includegraphics[width=0.95\linewidth]{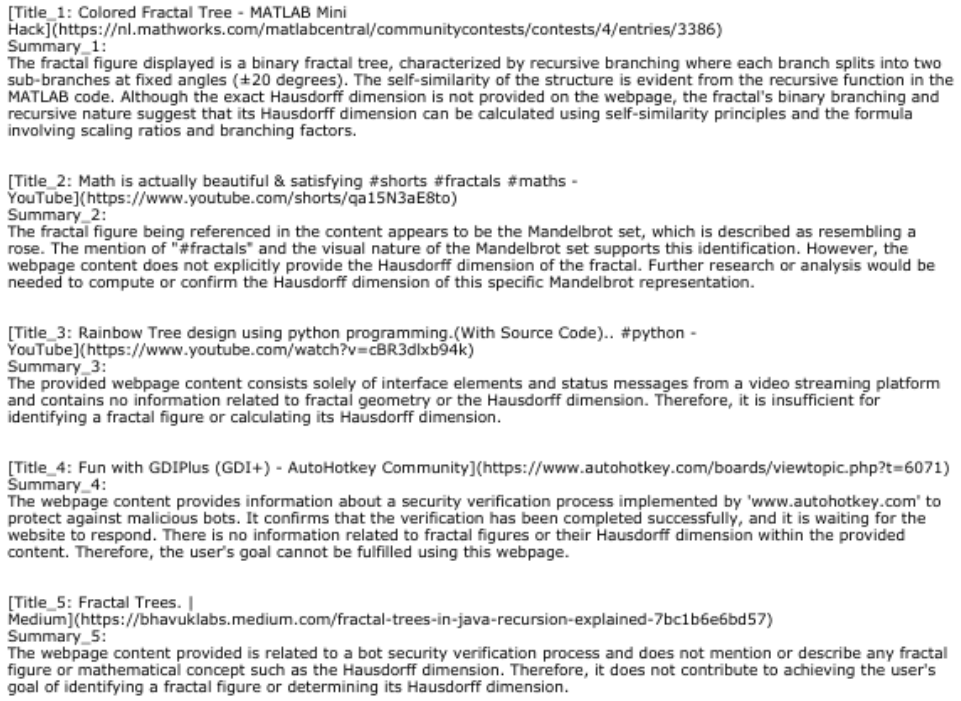}
\end{center}
\textbf{</tool\_response>}\\
\colorbox{gray!70}{%
    \parbox{\dimexpr\textwidth-2\fboxsep}{%
    \centering
        \textbf{Step 2}%
    }%
}\\
\textcolor{black}{\textbf{<think>} \\
The image search results indicate this is a "binary fractal tree" with recursive branching. However, the question mentions "removing the middle third from a line segment" which is characteristic of the Cantor set construction, not a binary fractal tree.
Let me reconsider. The question specifically mentions:
1. "Recursively removing the middle third from a line segment" - this is the Cantor set construction
2. "Listed by Hausdorff dimension" - this is a mathematical property
But the image shows a tree-like structure, not the Cantor set. There might be a mismatch between the question description and the actual figure, OR this could be a different fractal construction.
Let me search more specifically for fractals constructed by removing middle thirds and their Hausdorff dimensions.
\textbf{</think>}
}\\
\textcolor{black}{\textbf{<tool\_call>} \\ 
\{"name": "web\_search", "arguments": \{"queries": ["fractal removing middle third Hausdorff dimension", "Cantor set Hausdorff dimension", "middle third removal fractal dimension"]\}\}
\\
\textbf{</tool\_call>}}\\
\textcolor{black}{\textbf{<tool\_response>}\\
\begin{center}
  \includegraphics[width=0.95\linewidth]{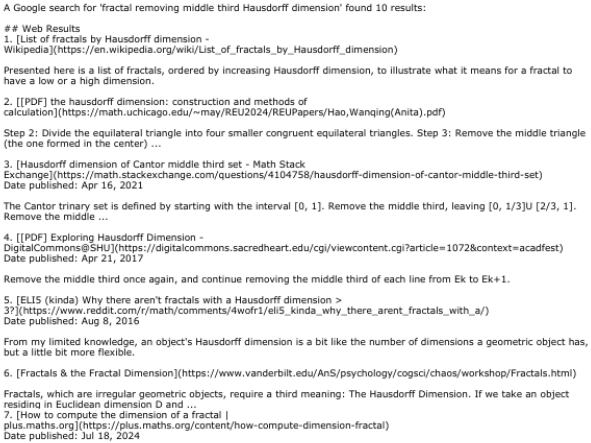}
\end{center}
\centering ... \\
\textbf{</tool\_response>}}\\
\colorbox{gray!70}{%
    \parbox{\dimexpr\textwidth-2\fboxsep}{%
    \centering
        \textbf{Step 3}%
    }%
}\\
\textcolor{black}{\textbf{<think>} \\
Perfect! Now I have the information I need. From the search results, I can see that:\\
1. The Cantor set is constructed by recursively removing the middle third from a line segment\\
2. The Hausdorff dimension of the Cantor set is log2/log3, which equals approximately 0.6309...\\
However, I need to verify if the image actually shows the Cantor set or a different fractal. The image search initially identified it as a "binary fractal tree," but the question specifically mentions "removing the middle third from a line segment."\\
Let me search more specifically to understand what fractal this might actually be, or if there's a connection between the Cantor set and tree-like structures. \\
\textbf{</think>}} \\
\textcolor{black}{\textbf{<tool\_call>} \\ 
\{"name": "web\_search", "arguments": \{"queries": ["Cantor set tree structure Hausdorff dimension", "fractal tree middle third removal dimension", "Cantor dust Hausdorff dimension"]\}\}
\\
\textbf{</tool\_call>}}\\
\textcolor{black}{\textbf{<tool\_response>}\\
\begin{center}
  \includegraphics[width=0.95\linewidth]{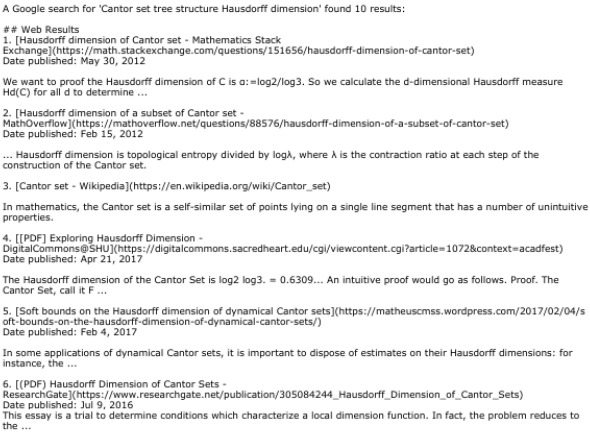}
\end{center}
\centering ...\\
\textbf{</tool\_response>}}\\
\colorbox{gray!70}{%
    \parbox{\dimexpr\textwidth-2\fboxsep}{%
    \centering
        \textbf{Step 4}%
    }%
}\\
\textcolor{black}{\textbf{<think>} \\
Now I have clear information. The Cantor set (and Cantor dust) has a Hausdorff dimension of log2/log3 $\approx$ 0.6309.\\
However, I still need to reconcile this with the image. The image search initially identified it as a "binary fractal tree" with recursive branching. But the question specifically mentions "removing the middle third from a line segment."\\
Let me think about this more carefully. There are a few possibilities:\\
1. The question might be misdescribing the figure - it could actually be a Cantor set or Cantor dust\\
2. The figure could be a tree-like representation of the Cantor set\\
3. It could be a different fractal construction\\
Looking at the search results, I found that "Cantor dust" is mentioned with a Hausdorff dimension of 1.2619 (which is 2×log2/log3). This is interesting - it's the Hausdorff dimension of the 2-dimensional Cantor set.\\
But the most fundamental Cantor set (1-dimensional) has Hausdorff dimension log2/log3 $\approx$ 0.6309.\\
Let me search more specifically to understand what this tree-like structure might represent and what its dimension is. \\
\textbf{</think>}} \\
\textcolor{black}{\textbf{<tool\_call>}\\
\{"name": "web\_search", "arguments": \{"queries": ["binary fractal tree Hausdorff dimension", "fractal tree recursive branching dimension formula", "Hausdorff dimension tree fractal construction"]\}\}
\textbf{</tool\_call>}}\\
\textcolor{black}{\textbf{<tool\_response>}\\
\begin{center}
  \includegraphics[width=0.95\linewidth]{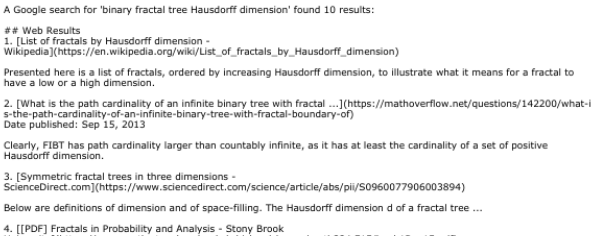}
\end{center}
\centering ...\\
\textbf{</tool\_response>}}\\
\colorbox{gray!70}{%
    \parbox{\dimexpr\textwidth-2\fboxsep}{%
    \centering
        \textbf{Step 5}%
    }%
}\\
\textcolor{black}{\textbf{<think>} \\
Now I have more information. From the search results, I can see that: 1. For fractal trees, the Hausdorff dimension is calculated using the formula D = log(N)/log(R), where: N = number of self-similar parts after each iteration. R = scaling factor (how much each branch is scaled down) 2. The image search identified this as a "binary fractal tree" with recursive branching. 3. The question mentions "removing the middle third from a line segment" which is characteristic of the Cantor set construction.\\
There seems to be a disconnect between the question description and the actual figure. Let me think about this more carefully. If this is a binary fractal tree (as identified by the image search), and if it's constructed by recursively removing the middle third, what would be its dimension? Actually, let me reconsider the question. It says: "constructed by recursively removing the middle third from a line segment and listed by Hausdorff dimension"\\
This could mean: - The figure is constructed using the middle third removal method (Cantor set construction) - OR it's a tree-like representation of such a construction\\
But the image clearly shows a tree structure, not the Cantor set. Let me search for information about whether there's a connection between middle third removal and tree structures, or if this might be a different fractal construction.\\
\textbf{</think>}} \\
\textcolor{black}{\textbf{<tool\_call>} \\ 
\{"name": "web\_search", "arguments": \{"queries": ["middle third removal fractal tree construction", "Cantor set tree visualization Hausdorff dimension", "fractal tree construction method middle third"]\}\}
\\
\textbf{</tool\_call>}\\
\textcolor{black}{\textbf{<tool\_response>}\\
\begin{center}
  \includegraphics[width=0.95\linewidth]{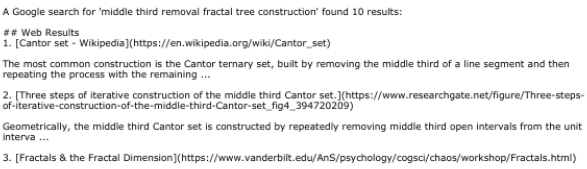}
\end{center}
\centering ...\\
\textbf{</tool\_response>}}}\\
\colorbox{gray!70}{%
    \parbox{\dimexpr\textwidth-2\fboxsep}{%
    \centering
        \textbf{Step 6}%
    }%
}\\
\textcolor{black}{\textbf{<think>} \\
Now I have a clearer picture. From the search results, I can see that:\\
1. The Cantor set is constructed by removing the middle third from a line segment\\
2. The Cantor set has Hausdorff dimension log2/log3 $\approx$ 0.6309\\
3. Fractal trees are constructed using different methods (like dividing into thirds and replacing the middle third with an equilateral triangle, or binary branching)\\
The key insight is that the question mentions "constructed by recursively removing the middle third from a line segment" - this is the Cantor set construction. However, the image shows a tree-like structure.\\
There are a few possibilities:\\
1. The question is misdescribing the figure\\
2. The figure is a tree-like representation of the Cantor set\\
3. It's a different fractal construction\\
But looking at the search results more carefully, I found this important information: "The Cantor ternary construction is a tree of nested subsets." This suggests that the Cantor set can be represented as a tree structure.\\
However, the most fundamental Cantor set (1-dimensional) has Hausdorff dimension log2/log3 $\approx$ 0.6309.\\
Let me think about what the question is really asking. It says: "What is the dimension of this figure, which is constructed by recursively removing the middle third from a line segment and listed by Hausdorff dimension?"\\
If this figure is indeed constructed by removing the middle third from a line segment (Cantor set construction), then the Hausdorff dimension is log2/log3.\\
But if it's a fractal tree constructed using a similar method (removing middle thirds and creating branches), then the dimension would be calculated differently.\\
Given that the image search identified it as a "binary fractal tree" and the question mentions "removing the middle third," I need to determine which interpretation is correct.\\
Actually, let me reconsider. The most straightforward answer to the question is that if a fractal is constructed by recursively removing the middle third from a line segment, it's the Cantor set, and its Hausdorff dimension is log2/log3.}\\
\textcolor{black}{
But the image doesn't look like the Cantor set - it looks like a tree. However, the question might be using "removing the middle third" metaphorically or the figure might be a different construction.\\
Let me search for the exact Hausdorff dimension of the Cantor set to be certain.\\
\textbf{</think>}} \\
\textcolor{black}{\textbf{<tool\_call>}\\
\{"name": "web\_search", "arguments": \{"queries": ["Cantor set Hausdorff dimension exact value log2/log3", "Hausdorff dimension of Cantor ternary set formula"]\}\}\\
\textbf{</tool\_call>}\\
\textcolor{black}{\textbf{<tool\_response>}\\
\begin{center}
  \includegraphics[width=0.95\linewidth]{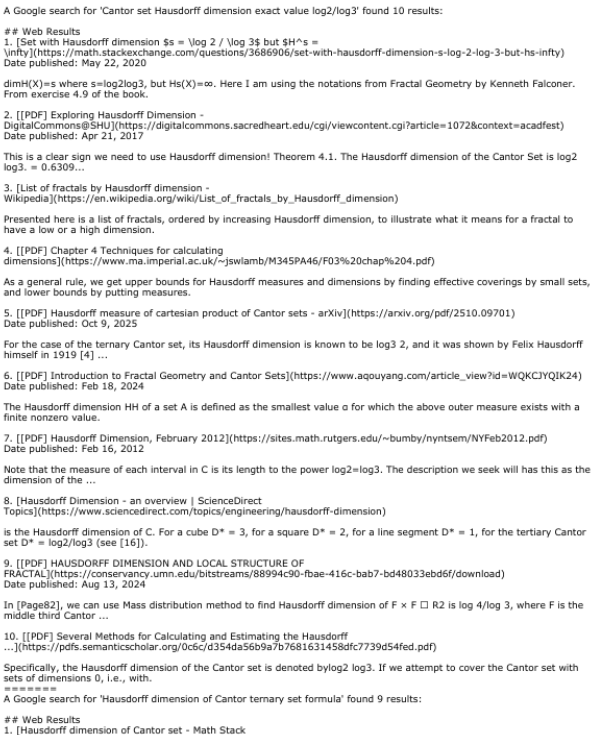}
\end{center}
\centering ...\\
\textbf{</tool\_response>}}}\\
\colorbox{gray!70}{%
    \parbox{\dimexpr\textwidth-2\fboxsep}{%
    \centering
        \textbf{Step 7}%
    }%
}\\
\textcolor{black}{\textbf{<think>} \\
Perfect! Now I have confirmed the exact Hausdorff dimension of the Cantor set. Multiple sources confirm that:\\
**The Hausdorff dimension of the Cantor set is log2/log3 $\approx$ 0.6309**\\
This is the standard result for the Cantor ternary set, which is constructed by recursively removing the middle third from a line segment.\\
However, I still need to reconcile this with the image. The image search identified it as a "binary fractal tree" with recursive branching. But the question specifically mentions "removing the middle third from a line segment."\\
There are a few possibilities:\\
1. The question is misdescribing the figure - it's actually the Cantor set\\
2. The figure is a tree-like representation of the Cantor set construction\\
3. It's a different fractal construction that uses a similar method\\
Given that the question explicitly states "constructed by recursively removing the middle third from a line segment," and this is the defining characteristic of the Cantor set, the answer should be the Hausdorff dimension of the Cantor set, which is log2/log3.\\
However, I should verify if there's any connection between the Cantor set and tree structures, or if this might be a different fractal altogether.\\
Let me think about this more carefully. The Cantor set is a 1-dimensional fractal. If the question is asking about a tree-like structure constructed using the middle third removal method, it could be:\\
- A 2-dimensional Cantor set (Cantor dust) which has Hausdorff dimension 2×log2/log3 $\approx$ 1.2619\\
- Or a different construction\\
But the most straightforward interpretation is that the question is referring to the Cantor set construction, and the Hausdorff dimension is log2/log3.\\
Let me do one more search to see if there's any specific connection between this tree structure and the Cantor set.\\
\textbf{</think>}} \\
\textcolor{black}{\textbf{<tool\_call>}\\
\{"name": "web\_search", "arguments": \{"queries": ["Cantor set tree structure visualization", "ternary Cantor set tree representation", "Cantor set as tree fractal"]\}\}
\textbf{</tool\_call>}\\
\textcolor{black}{\textbf{<tool\_response>}\\
A Google search for 'Cantor set tree structure visualization' found 10 results:\\
\#\# Web Results\\
1. [Cantor Set - Wolfram Demonstrations Project](https://demonstrations\\.wolfram.com/CantorSet/)\\
The Cantor set is obtained from[0,1] by removing the open middle third and iteratively removing the middle third of each remaining interval.\\
2. [[PDF] Cantor sets, binary trees, and Lipschitz ...]\\
...\\
\textbf{</tool\_response>}}}\\
\colorbox{gray!70}{%
    \parbox{\dimexpr\textwidth-2\fboxsep}{%
    \centering
        \textbf{Step 8}%
    }%
}\\
\textcolor{black}{\textbf{<think>} \\
Excellent! Now I found the key information.\\ From the search results, I can see:\\
1. From result \#5 in the first search: "Often trees are represented by joining any element to each of its immediate successors. Figure 1 shows part of the Cantor tree, where the horizontal segments."\\
2. From result \#5 in the third search: "The Cantor Set is one of the simplest fractals, and also one of the earliest to have been studied in detail. As a one-dimensional structure, it was easy to..."\\
3. From result \#1 in the third search: "One of the first examples of a fractal is the Cantor Set, which was discovered by Georg Cantor in 1883."\\
The key insight is that the Cantor set can be represented as a tree structure. The Cantor tree is a tree where each node has two children, representing the binary choices in the ternary (base-3) representation of numbers in the Cantor set.\\
So the image appears to be showing the Cantor tree - a tree structure representing the Cantor set. The "removing the middle third" construction is the underlying process that generates this tree structure.\\
Therefore, the Hausdorff dimension of this figure (the Cantor tree, which represents the Cantor set) is the Hausdorff dimension of the Cantor set, which is:\\
**log2/log3 $\approx$ 0.6309**\\
This is the standard result for the Cantor ternary set.\\
\textbf{</think>}} \\
\textcolor{black}{\textbf{<answer>}\\
Based on my research, I can now provide a comprehensive answer to this question.\\
The figure shown is actually the **Cantor tree**, which represents the Cantor set construction. The Cantor set is constructed by recursively removing the middle third from a line segment - starting with [0,1], then removing the middle third (1/3, 2/3), then removing the middle third of each remaining interval, and so on.\\
The Cantor set can be represented as a tree structure where:\\
- The root represents the interval [0,1]\\
- Each node branches into two children representing the left and right thirds of the interval\\
- This creates a binary tree structure where each level represents one iteration of the construction\\
The Hausdorff dimension of the Cantor set (and therefore of the Cantor tree representing it) is calculated using the formula for self-similar fractals:\\
**D = log(N)/log(R)**\\
Where:\\
- N = number of self-similar parts after each iteration = 2 (each interval splits into 2 parts)\\
- R = scaling factor = 1/3 (each branch is 1/3 the length of the parent)\\
Therefore:\\
**D = log2/log3 $\approx$ 0.6309**\\
This is the well-known Hausdorff dimension of the Cantor ternary set.\\
\textbackslash boxed\{$\log_2 / \log_3$\} \textcolor{black}{\checkmark} \\
\textbf{</answer>}}
\end{tcolorbox}